\documentclass[runningheads]{llncs}
\usepackage{graphicx}

\usepackage{tikz}
\usepackage{comment}
\usepackage{amsmath,amssymb} 
\usepackage{color}
\usepackage{amsmath,graphicx}
\usepackage{booktabs}
\usepackage{multirow}
\usepackage{caption}
\usepackage{subcaption}
\usepackage{amsmath}
\usepackage{amssymb}
\usepackage{comment}
\usepackage{float}
\usepackage{booktabs}
\usepackage{cite}
\usepackage[T1]{fontenc}
\usepackage{hyperref}

\usepackage{array}
\usepackage{multirow}
\usepackage{textcomp}
\usepackage{booktabs}
\usepackage{graphicx}
\usepackage{tabularx}
\usepackage{tabulary}
\usepackage{pdfpages}
\usepackage{tikz}
\usepackage{tablefootnote}
\usepackage{textcomp}
\usepackage{verbatim}
\usepackage{makecell}
\usepackage{enumitem}
\usepackage{comment}
\usepackage{subcaption}
\usepackage[T1]{fontenc}
\usepackage{lscape}
\usepackage[list=true]{subcaption}
\usepackage{hyperref}
\usepackage{textcomp}
\usepackage[flushleft]{threeparttable}
\usepackage{float} 

\begin{document}
\setcounter{secnumdepth}{3}

\pagestyle{headings}
\mainmatter
\def\ECCVSubNumber{100}  

\title{Computer Vision for Clinical Gait Analysis: \\ A Gait Abnormality Video Dataset} 

\titlerunning{Computer Vision for Clinical Gait Analysis: \\ A Gait Abnormality Video Dataset}

\author{Rahm Ranjan\inst{1,2} \and
David Ahmedt-Aristizabal\inst{1} 
\and
Mohammad Ali Armin\inst{1}
\and
Juno Kim\inst{2}
}
\authorrunning{Ranjan et al.}
%
\institute{Imaging and Computer Vision Group, CSIRO Data61, Canberra, Australia 
\email{\{david.ahmedtaristizabal,ali.armin\}@data61.csiro.au}
\and
University of New South Wales, Sydney, Australia 
\email{\{rahm.ranjan,juno.kim\}@unsw.edu.au}
}
\maketitle

\begin{abstract}
\vspace{-2pt}
Clinical gait analysis (CGA) using computer vision is an emerging field in artificial intelligence that faces barriers of accessible, real-world data, and clear task objectives. 
This paper lays the foundation for current developments in CGA as well as vision-based methods and datasets suitable for gait analysis. 
We introduce The Gait Abnormality in Video Dataset (GAVD) in response to our review of over 150 current gait-related computer vision datasets, which highlighted the need for a large and accessible gait dataset clinically annotated for CGA. 
GAVD stands out as the largest video gait dataset, comprising 1874 sequences of normal, abnormal and pathological gaits. Additionally, GAVD includes clinically annotated RGB data sourced from publicly available content on online platforms. It also encompasses over 400 subjects who have undergone clinical grade visual screening to represent a diverse range of abnormal gait patterns, captured in various settings, including hospital clinics and urban uncontrolled outdoor environments. We demonstrate the validity of the dataset and utility of action recognition models for CGA using pretrained models Temporal Segment Networks(TSN) and SlowFast network to achieve video abnormality detection of 94\% and 92\% respectively when tested on GAVD dataset. A GitHub repository \href{https://github.com/Rahmyyy/GAVD}{GAVD} consisting of convenient URL links, and clinically relevant annotation for CGA is provided for over 450 online videos, featuring diverse subjects performing a range of normal, pathological, and abnormal gait patterns. 

\vspace{-5pt}
\end{abstract}

\section{Introduction}

\label{sec:intro}

Clinical gait analysis (CGA) is a well-established method of understanding and monitoring human health and performance by identifying abnormalities in gait pattern movements \cite{Baker2016Gait}. 
Traditionally, the interpretation of human movement relies on clinical experience and observation skills to identify abnormalities and provide explanations for deviations in movement patterns. This conventional approach is heavily dependent on the subjective judgement of experts that can vary between clinicians. It also poses challenges in objectively measuring movement without the availability of resource-intensive and expensive motion capture equipment, which is often not available in many clinical settings. Further to this, tools for patients to monitor their movement patterns with regards to disease progression or rehabilitation are also limited thus increasing the burden on healthcare to monitor movements that could otherwise be self-managed with objective feedback. 

With notable advancements in computer vision and machine learning, video-based human action recognition has emerged as a powerful tool for gait analysis, providing clinicians with enhanced usability and utility~\cite{sethi2022}. Current computer vision research in gait focused on action classification\cite{kong2022}, human gait identification~\cite{Sepas-Moghaddam2023} and gait pathology classification \cite{sethi2022}. While these tasks are important, they fall short in providing clinically relevant information, particularly regarding the spatial and temporal location of movement abnormalities\cite{roberts2017biomechanical}. 
Furthermore, existing technologies such as 3D motion capture do not allow health and medical specialists to retrospectively use data formats like monocular RGB video, the most prevalent method for capturing clinical movement for analysis~\cite{hensley2020video}. 
Our suspicion is that the limited progress in computer vision research for the task of localising spatial and temporal abnormalities for CGA, is attributable, in part, to 1) a lack of understanding regarding the clinical requirements for this information and 2) the limited availability of clinically annotated video data essential for training and testing models. 

Many human action recognition tasks typically involve the classification of course action classes, such as cycling and yoga \cite{le2022comprehensive}. However, intra-class variations within an action is crucial for recognising what is considered typical or atypical for a given activity \cite{Parashar2022Intraclass}. A finer level of action analysis, in terms of both spatial and temporal information, provides an understanding of subtle differences between actions\cite{Shao2020FineGym}. This information also allows clinical practitioners to target and instruct patients to modify movements to improve the speed, quality or efficiency of an action while avoiding detrimental movements that can cause injury. 

To date, current state-of-the-art methods in gait analysis employing deep learning have shown great promise~\cite{kidzinski2020deep}; however, their integration into clinical practice remains challenging. These methods often rely on depth sensors, inertial measurement units (IMUs), or 3D motion capture systems which are often inaccessible or impractical in many clinical scenarios such as monitoring a person in their home environment. Furthermore, existing gait analysis techniques tend to be narrowly focused, concentrating on a limited set of gait types and pathologies. As a result, these models encounter challenges in achieving generalisability across a diverse range of gait and movement abnormalities. 
In addition to these limitations, current methods primarily focus only on general classification without providing explainability via spatial or temporal location information. Sethi et al.~\cite{sethi2022} provides a comprehensive overview of the current state of action recognition research and the challenges associated with gait classification. However, we extend the scope of clinical gait analysis research towards recent computer vision approaches such as the use of action recognition to identify abnormal gait patterns from monocular video data.

\subsection{Contributions}  
\begin{figure}
    \centering
    \includegraphics[width=1\linewidth]{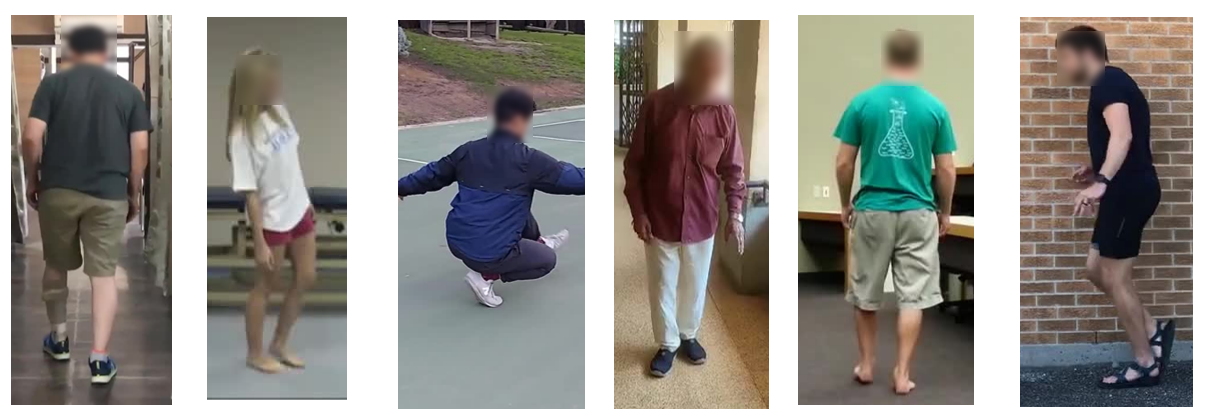}
    \caption{Example frames representing a variety of subjects from GAVD dataset}
    \label{img:GAVD_examples}
    \vspace{-4pt}
\end{figure}

Despite the availability of numerous gait datasets such as the CASIA \cite{Wang2003Silhouette} and OU-ISIR MVLP \cite{Takemura2018Multiview} datasets, there is a lack of abnormal gait datasets that can be applied to a variety of gait analysis tasks, especially in an in-the-wild clinical setting. To address this gap, we present the following contributions.
\begin{enumerate}
\vspace{-4pt}
    \item A comprehensive summary encompassing over 150 existing datasets related to gait for computer vision with analysis of key features such as the number of subjects, data modalities, gait type and availability of RGB data are included.

    \item The Gait Abnormality in Video Dataset (GAVD), an accessible video dataset annotated by clinical professionals to support future development of computer vision methods for clinically validated gait analysis (See Figure~\ref{img:GAVD_examples}). 
    GAVD aims to improve the state of the art in gait analysis and gait classification research while overcoming barriers such as lack of medically validated movement data. We focus on video data as this will provide utility for trending healthcare developments such as telehealth \cite{koonin2020trends} and self-monitoring for healthcare \cite{mcbain2015impact}. 

    \item Baseline results investigating the suitability of region-based action recognition models for the task of gait abnormality classification. 

\end{enumerate}

\section{Background}

\subsection{Clinical gait analysis}

Clinical gait analysis (CGA) is an important diagnostic tool employed in clinical settings to assess various musculoskeletal and neurological conditions that impact normal human movement~\cite{Wren2020Clinical}. Trained experts in human movement, such as physiotherapists and medical specialists, systematically measure, evaluate, and interpret multiple gait parameters. These parameters provide information into potential abnormalities or impairments~\cite{baker2006gait}. 

The quantitative and qualitative data acquired through clinical gait analysis encompass kinematics, kinetics, and muscle activity during walking. This information enables healthcare professionals to develop targeted treatment plans and ongoing monitoring of progress~\cite{whittle2014gait}. 

The primary objective of clinical gait analysis is to identify deviations from typical gait patterns and understand the mechanisms causing deviations. Many internal and external factors contribute to abnormalities in gait~\cite{chang2010role}. External factors such as the environment, terrain, clothing, and carried items have great influences on an individual's gait pattern~\cite{baker2006gait}. However, for most clinical gait analyses, there is a focus on internal factors such as range of motion, muscle strength, and flexibility~\cite{whittle2014gait} as well as neurological~\cite{jahn2010gait} and psychological factors~\cite{lucyyardley2004psychosocial}. 

Clinical references for typical human movement are used to classify gait normality in the clinical setting~\cite{roberts2017biomechanical}. Unlike a singular universal metric, clinicians describe movement using many different gait parameters, each associated with specific reference norms tailored for various purposes and conditions. For example, clinicians may recognise a festinating or shuffle gait typically exhibited by individuals with Parkinson's Disease.  Abnormalities in joint angles, stride length, step width, and other parameters are systematically observed to aid clinical decision-making~\cite{Stergiou2020Biomechanics}.

Deviations from typical ranges of movement can indicate underlying conditions such as muscle weakness~\cite{sicard2002gait}, joint stiffness~\cite{totah2019impact}, balance issues~\cite{baker2006gait}, or neurological disorders~\cite{jahn2010gait}. The analysis can also identify compensatory movements that patients may adapt to accommodate their impairments~\cite{whittle2014gait}, providing valuable insights into the functional limitations and potential injury risks. The location and timing of the abnormal movement provide much information regarding the underlying nature of the issue and potential solutions~\cite{baker2006gait}. This information guides the development of personalised treatment plans, which may involve rehabilitation exercises, orthotic devices, surgical interventions, or other interventions tailored to address the specific impairments identified.

Typically, objective and accurate clinical gait analysis requires specialised equipment to capture motion and force~\cite{wren2011efficacy}. Sensors, optical devices, force plates, electromyography (EMG), and other instruments are used to capture and measure various aspects of gait. These technologies enable the collection of accurate and objective data, allowing for quantitative assessment and comparison of gait parameters across different individuals and time points~\cite{di2020gait}.

Access to specialised equipment for clinical gait analysis is often challenging for many healthcare professionals due to various factors such as cost, and geographical constraints~\cite{krebs1985reliability}. The high cost of sophisticated gait analysis systems, including marker-based motion capture systems, force plates, and electromyography (EMG) devices, can pose a significant barrier for smaller clinics or healthcare facilities with limited budgets. Furthermore, these specialised instruments require technical expertise for setup, calibration, and data interpretation, which may not be readily available to all clinicians~\cite{krebs1985reliability}. 

As a result, the best practice for many clinicians is to use alternative methods and tools to assess gait and obtain valuable information about a patient's walking patterns~\cite{roberts2017biomechanical}. Typically, these methods of assessment are cost effective and require minimal equipment. However, there is a great trade-off between usability, accuracy and objectivity. As tools rely more on clinical experience and observation, their ability to be objective decreases. Alternative gait analysis tools include:

\vspace{4pt}
\textit{Visual observation:}
Clinical gait assessments often begin with a visual observation of the patient's walking patterns. Skilled clinicians can detect noticeable abnormalities or compensatory movements by carefully observing the patient's gait, posture, and body mechanics during walking. Visual observation provides valuable qualitative information that can guide initial treatment decisions~\cite{whittle2014gait}. However, it is difficult to achieve consistent and accurate data collection given the subjective nature of this assessment~\cite{baker2006gait}. 

\vspace{4pt}
\textit{Functional assessments:}
Functional assessments involve evaluating a patient's performance in functional tasks related to gait, such as the Berg balance tests~\cite{bogle1996use}, or the 6-minute walk~\cite{bittner1993prediction} tests. These assessments focus on specific aspects of gait function and can provide insights into overall mobility, endurance, and functional limitations. However, they are limited in providing information regarding timing and location of movement issues.

\vspace{4pt}
\textit{Questionnaires and self-reports:}
Standardised questionnaires and self-report measures are used to gather subjective information about a patient's gait-related symptoms, perceived functional limitations, and impacts on daily activities \cite{Brach2002Identifying}. These tools help in assessing the patient's perspective and can complement objective gait analysis data.

\vspace{4pt}
\textit{Portable sensors and wearable devices:}
Technological advancements have led to the development of cost-effective, portable sensors and wearable devices that can simply capture skeletal motion data. Examples include accelerometers, gyroscopes, and inertial measurement units (IMUs) that can be attached to the body or integrated into wearable devices. This has the advantage of capturing long term data collection and use in a variety of environments~\cite{benson2022real}. However, Sensor-based motion capture alternatives do not provide the same level of accuracy as optical based equipment and are difficult to place accurately on wearers~\cite{mengucc2014wearable}

\vspace{4pt}
\textit{Video-based analysis:}
Video recordings of a patient's gait serve as a useful tool for qualitative analysis. Clinicians can review the recorded videos frame-by-frame to identify gait abnormalities, joint angles, and temporal parameters. Video capture assists with clinical human movement analysis and is considered best clinical practice~\cite{wiles2003use}, given the availability of integrated cameras and storage in mobile computing devices. Video-based analysis has been validated as a less accurate but clinically useful alternative to more expensive marker-based optical motion capture systems, with the added benefit of usability outside of the clinical environment~\cite{Nakano2020Evaluation}.

\subsection{Overview of human action recognition in videos}
Human action recognition is the computer vision task of understanding the actions of humans based on visual appearances, motion and context~\cite{Camarena2023Concise}. This broad task has many successful methods that have utilised both appearance-based and model-based approaches \cite{Ahmedt-Aristizabal2024Deep}. The main objective of human action recognition is to classify human actions in video. Initially, many approaches were based on the success of CNN in image classification, which was able to extract spatial features. However, the sequential nature of actions in video requires both temporal and spatial information. The focus on combining temporal and spatial features to recognise actions mimics the human biological visual process and was popularised by~\cite{Simonyan2014TwoStream}. The Temporal Segment Network (TSN) proposed by~\cite{Wang2016Temporal} provided a framework addressing the issue of long-range temporal structure modelling. This model utilises four available modalities from RGB data including the single RGB image, stacked RGB difference, stacked optical flow field, and stacked warped optical flow field, making it possible to train deep learning networks using limited training data without severe over-fitting. 

The availability of large video datasets for action recognition such as Kinetics-700~\cite{Kay2017Kinetics} and computer vision challenges such as the Activity Net \cite{Heilbron_2015_CVPR} spurred research towards deep learning methods. State of the art methods such as \cite{Gammulle2017Two} utilised recurrent neural network (RNN) methods such as long short-term memory (LSTM) networks which have successfully been used with other sequential data domains such as audio and signal classifications. \cite{Feichtenhofer2019SlowFast} identifies the value of slow changes in categorical semantics while fast changes in the motion being performed for an action. This led to the development of two-pathway methods such as SlowFast, which can detect changes in motion without the need for optical flow calculations.

\subsection{Vision-based approaches for gait recognition}
Computer vision has been well-established for the tasks of gait recognition, person identification and human action recognition. These tasks are grounded in surveillance and biometrics\cite{Harris2022Survey}. Gait recognition is the task of using gait as a biometric feature to identify a person based on their body shape, posture, and walking pattern. Similar to other biometrics such as fingerprints, a gait pattern is unique to an individual and has the advantage of being a characteristic identify-able at a distance\cite{FilipiGoncalvesDosSantos2023Gait}. Gait recognition is primarily concerned with person identification, however, there are sub-tasks such as person re-identification\cite{Harris2022Survey}, emotion detection~\cite{sheng2021multi,Bhattacharya2020STEP}, age estimation\cite{Xu2019Gaitbased}, and gender recognition \cite{Zhang2013Estimation}. 

Given the limited approaches using computer vision for the task of clinical gait analysis using only RGB video data, gait recognition methods offer many techniques that are transferable for aspects of gait analysis. Gait features extracted in gait recognition can be used as important features to identify and classify gait abnormalities~\cite{Kim2022deeplearning}. 

Approaches for gait recognition are categorised into appearance-based or model-based methods. Appearance-based methods utilise extracted features of gait sequences represented by silhouette images, such as the gait energy image (GEI) \cite{Han2006Individual}. Model-based methods fit models such as human pose estimation skeletons to the gait sequence images \cite{Loureiro2020Using}. The main advantage cited by proponents of model-based approaches is that motion information is separated from appearance which can be biased by covariates such as camera viewing angle, clothing, and accessories.

\vspace{4pt}
\textit{Appearance-Based:}
GEINet\cite{Shiraga2016GEINet} is a method for gait recognition utilising CNN to extract and learn gait features from GEIs using the cross-entropy loss. 
GaitSet \cite{Chao2019GaitSet} uses a larger CNN to cross-match views that are input as silhouettes from a sequence of gait. The varying position of the person relative to the frame can be considered as changing granularity that is a feature of the model. 
GaitPart \cite{Fan2020GaitPart} introduces a temporal part-based model for gait recognition. This learns part-level features and models short-range temporal patterns using a novel Micro-motion Capture Module which are then aggregated as the output feature. 
MT3D \cite{Lin2020Gait} proposes a new framework using multiple-temporal-scale that integrates temporal information from multiple temporal scales to fuse both frame and interval information. Here, a 3D CNN is used to enhance the spatial-temporal modelling capability. 
REALgait \cite{Zhang2023RealGait} attempts to improve the performance of gait recognition in uncontrolled environments and establish the benefit of combining both appearance and motion information for gait recognition. This approach highlights that the body silhouette input, which is prevalent in many appearance-based gait recognition methods, is severely corrupted or influenced by external factors, such as clothing and accessories.

\vspace{4pt}
\textit{Model-Based:}
PoseGait \cite{Liao2020modelbased}. Proposes to overcome viewpoint issues for gait recognition by extracting spatiotemporal features from 3D pose information using CNN and LSTM. They use 2D pose to infer the 3D pose and then extract features including pose, joint angles, limb lengths, and motion to improve the gait authentication rate. 
Gait D \cite{Gao2022GaitD} take a skeleton‐based gait recognition algorithm approach to solve the problem of covariate conditions. The spatial and temporal relationships of joints are used as feature inputs into the model. Insignificant features are removed using a feature map to achieve a better recognition rate in the presence of covariate factors. 
Gait Graph II \cite{Teepe2022Deeper} proposes a GCN architecture that combines higher-order inputs, and residual networks to create an efficient architecture for gait recognition. \cite{Yousef2023Modelbased} introduces a fusion approach combining both model and appearance-based methods for person identification.

\subsection{Vision-based approaches for clinical gait analysis}
Computer vision methods are increasingly utilised in health care due to their cost-effectiveness, user-friendliness, objectivity, and automation capabilities~\cite{Milstein2020Computer}. For gait analysis, these techniques have been instrumental in extracting gait parameters for classifying pathologies such as Parkinson's Disease, Huntington's disease, cerebral palsy, and stroke.
While there is a growing body or research in this area, there is limited use of appearance-based and hybrid appearance with pose methods for gait analysis. The majority of methods relying on pose-based approaches utilise skeletal data from depth sensor cameras (such the Microsoft Kinect) or 3D motion capture setup. This preference is likely due to the limited availability of video datasets for abnormal gait patterns, which is discussed in further detail below.  
For instance, Kidzinski et al, \cite{kidzinski2019automatic} trained an RNN on hospital-owned gait data to detect gait events in children with cerebral palsy. However, their dataset is not publicly available and is specifically limited to children with cerebral palsy. The authors extend this study~\cite{kidzinski2020deep} by demonstrating the feasibility of kinematic analysis using monocular RGB video through pose estimation and CNNs.  

The use of  RNNs is popular, as demonstrated by \cite{Khokhlova2019Normal} who trained an LSTM with skeletal data from RGB-D sensors to detect gait abnormality. They introduced the Multi-modal gait symmetry (MMGS) dataset in response to the limited availability of abnormal gait patterns in publicly available datasets. However, the MMGS is limited in its representation of pathological gaits and lacks video data containing only simulated gait motions from 27 actors. 
\cite{Kim2022deeplearning} emphasised the importance of accurate identification of the gait cycle and events for gait analysis using 3D motion tracking data. They employed LSTM to identify fine-grained gait events in the gait cycle.
\cite{Stenum2021Twodimensional} demonstrated the ability to estimate spatiotemporal gait parameters as well as hip and knee angles accurately using using pose estimation. However, this approach is constrained by the participant's position relative to the camera, which can influence spatial measures such as step length. Hence, there is a growing need for view-invariant methods to improve clinical gait analysis. 
\cite{Tian2022Skeletonbased} proposes a spatiotemporal attention-enhanced gait-structural graph convolutional network (AGS-GCN) that uses lower-limb pose estimation to extract skeletal data to classify vestibular pathologies from gait. Similarly, \cite{Kim2022PathologicalGait} employed a spatiotemporal graph convolutional network model (ST-GCN) to classify pathological gaits using inputs derived from preprocessing data from RGB-D skeletal data.  
\cite{Jun2023Hybrid} fused features from GCNs, RNNs, and basic gait parameters such as step length to classify various neurological pathologies from gait. 
\cite{cheriet2023multi} move away from CNN networks and introduced a Multi-Speed transformer technique that utilises skeletal coordinates from pose estimation to classify gait patterns. They provided the Beside Gait dataset, which offers a large representation of abnormal gait sequences but lacks RGB video data.


\section{Datasets}
Gait datasets are common in computer vision for tasks such as gait identification, however, there is a great need for clinically suitable gait data that provides accessible video data for tasks such as gait abnormality identification.

\subsection{Overview of current gait datasets for computer vision}
In this section, we provide a comprehensive overview and summary of existing gait datasets and their features. While numerous video datasets have been cited in published literature for gait-related activities in computer vision, there is a notable absence of clinically appropriate, sizable, and available datasets specifically tailored for clinical gait analysis using RGB data. This gap highlights the need for a new gait dataset, namely the Gait Abnormality in Video Dataset (GAVD), that overcomes the limitations of current datasets to enable improved clinical gait analysis. Detailed information about GAVD is provided later in this paper.  

\begin{figure}
    \centering
    \includegraphics[width=1\linewidth]{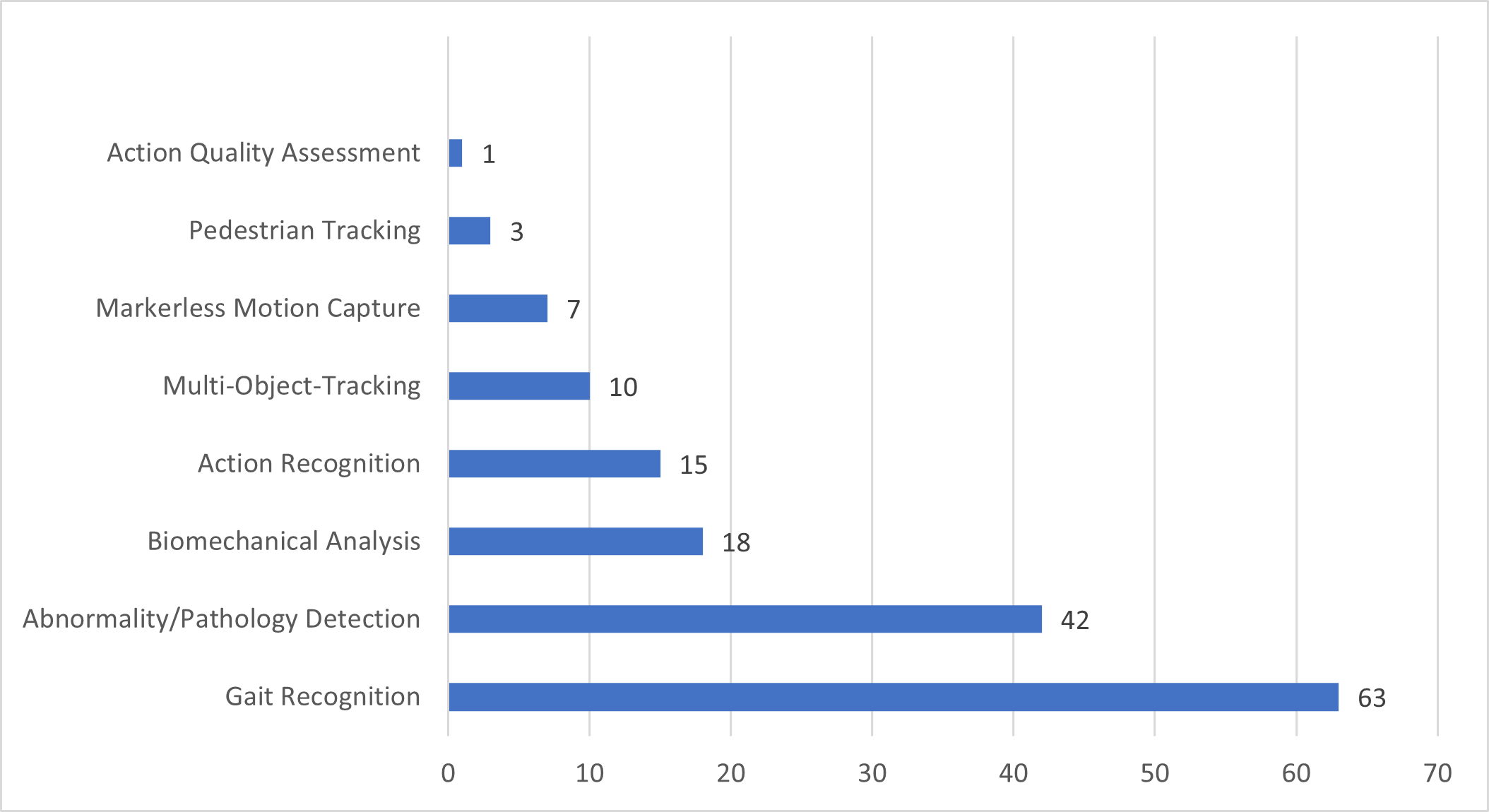}
    \caption{Distribution of Gait Datasets by Computer Vision Task}
    \label{fig:number_datasets}
    \vspace{-4pt}
\end{figure}

\begin{table*}[!t]
\caption{Datasets with Normal Gait Actions and Activities}
\vspace{-1pt}
\centering
\label{table:Normal-datasets}
\resizebox{0.99\textwidth}{!}{%
\begin{tabular}{l|l c c c l l c}
\toprule
\# & \textbf{Dataset (year)}  & \textbf{Total Subjects} & \textbf{Total Gait Sequences} & \textbf{Original Purpose} & \textbf{Data Modalities} & \textbf{Gait Type} & \textbf{RGB Available}\\ 
\midrule
1 &DAI(2016)\cite{Nieto-Hidalgo2016vision}&2&30&AP&RGB, Sil&Norm&No\\
2 &Ismail(2017)\cite{Ismail2017Torwads}&11&N/S&AP&3D-S&Norm&No\\
3 &KTH (Class - Walking)(2004)\cite{Schuldt2004Recognizing}&25&N/S&AR&RGB&Norm&No\\
4 &IXMAS (Class - Walk)(2006)\cite{Weinland2006Free}&10&50&AR&RGB&Norm&No\\
5 &Ogale (Class - Walk)(2007)\cite{Ogale2007ViewInvariant}&10&80&AR&RGB&Norm&No\\
6 &Weizmann(Class - Walk)(2007)\cite{Gorelick2007Actions}&9&9&AR&RGB&Norm&No\\
7 &Huang(Class - Walk)(2008)\cite{Huang2008Action}&8&48&AR&RGB&Norm&No\\
8 &i3DPost (Class - Walk)(2009)\cite{Gkalelis2009i3DPost}&8&64&AR&RGB&Norm&No\\
9 &HumanEva I (Walking Class)(2010)\cite{Sigal2010HumanEva}&4&12&AR&3D Mocap, RGB, &Norm&No\\
10 &MuHAVi(2010)\cite{Singh2010MuHAVi}&14&112&AR&Sil&Norm&No\\
11 &HMDB51(Class - Walk)(2013)\cite{Kuehne2013HMDB51}&N/S&100(approx.)&AR&RGB&Norm&Yes\\
12 &PKU-MMD(Class - Walking)(2017)\cite{Liu2017PKUMMD}&66&N/S&AR&3D-S, depth, RGB&Norm&No\\
13 &N-UCLA(Class - Walk)(2014)\cite{Wang2014Crossview}&10&N/S&AR &Depth, RGB, 3D-S&Norm&Yes\\
14 &NTU RGB+D (Class - Walk)(2016)\cite{Shahroudy2016NTU}&N/S&N/S&AR &3D-S, Depth, RGB, TIR&Norm&Yes\\
15 &NTU RGB+D 120 (Class - Walk)(2016)\cite{Liu2020NTU}&N/S&N/S&AR &3D-S, Depth, RGB, TIR&Norm&Yes\\
16 &UWA3D Multiview II(2016)\cite{Rahmani2016Histogram}&10&N/S&AR &Depth, RGB, 3D-S&Norm&Yes\\
17 &Indonesian Gait Dataset(2012)\cite{Mahyuddin2012Development}&212&212&BA&2D-S&Norm&No\\
18 &3D Indonesian Gait(2015)\cite{Anggraeni2015Gait}&60&60&BA&3D Mocap (Lower Limb)&Norm&No\\
19 &Human Gait Phase(2015)\cite{Hebenstreit2015Effect}&21&252&BA&3D Mocap (Lower Limb)&Norm&No\\
20 &Fukuchi(2018)\cite{Fukuchi2018public}&42&1409&BA&3D Mocap (Lower Limb)&Norm&No\\
21 &Rocha(2018)\cite{Rocha2018System}&20&20&BA&3D Mocap, 3D-S&Norm&No\\
22 &Geneva Asymptomatic Gait(2023)\cite{Grouvel2023dataset}&10&337&BA&3D Mocap&Norm&No\\
23 &Topham(2023)\cite{Topham2023diverse}&64&3120&BA&3D-S, RGB &Norm&Yes\\
24 &UCSD ID(1998)\cite{Little1998Recognizing}&6&42&GR&OF&Norm&No\\
25 &SOTON(1999)\cite{Nixon1999Automatic}&5&26&GR&RGB, Sil&Norm&Yes\\
26 &CMU MoBo(2001)\cite{Gross2001CMU}&25&600&GR&RGB, Sil&Norm&No\\
27 &Johnson(2001)\cite{Johnson2001multiview}&18&N/S&GR&Sil&Norm&No\\
28 &MIT(2002)\cite{Collins2002Silhouettebased}&25&194&GR&Sil&Norm&No\\
29 &CASIA - A(2003)\cite{Wang2003Silhouette}&20&240&GR&RGB, Sil&Norm&Yes\\
30 &Georgia Tech Gait(2004)\cite{Tanawongsuwan2004Modelling}&6&4&GR&RGB, Sil&Norm&No\\
31 &Southampton-HID(2004)\cite{Shutler2004Large}&100(approx.)&800(approx.)&GR&RGB, Sil&Norm&Yes\\
32 &HumanID(2005)\cite{Sarkar2005humanID}&122&1870&GR&RGB&Norm&No\\
33 &CASIA - C(2006)\cite{Tan2006Efficient}&153&1530&GR&Sil, TIR&Norm&No\\
34 &KY4D(2010)\cite{Iwashita2010Person}&41&N/S&GR&RGB&Norm&No\\
35 &PRID2011(2011)\cite{Hirzer2011Person}&200&N/S&GR&RGB&Norm&Yes\\
36 &TUM-IITKGP(2011)\cite{Hofmann2011Gait}&35&840&GR&GEI&Norm&No\\
37 & 3D gait volume dataset(2012)\cite{Muramatsu2012Arbitrary}&72&1728&GR&3D Mesh, RGB, Sil&Norm&No\\
38 &DGait(2012)\cite{Borras2012Depth}&53&N/S&GR&3D-S&Norm&No\\
39 &OU-ISIR - (Large Population)(2012)\cite{Iwama2012OUISIR}&4007&N/S&GR&Sil&Norm&No\\
40 &OU-ISIR - Treadmill(2012)\cite{Makihara2012OUISIR}&487&8728&GR&Sil&Norm&No\\
41 &SOTON Temporal(2012)\cite{Matovski2010effect}&25&2000(approx.)&GR&GEI, RGB, Sil&Norm&No\\
42 &BUAA-IRIP(2013)\cite{Zhang2013Estimation}&86&3010&GR&GEI&Norm&No\\
43 &AVA(2014)\cite{Lopez-Fernandez2014AVA}&20&1200&GR&RGB&Norm&No\\
44 &Kinect Gait Biometry(2014)\cite{Araujo2014Kinect}&164&820&GR&3D-S&Norm&No\\
45 &KIST(2014)\cite{Yun2014Statistical}&113&113&GR&3D Mocap (Lower Limb)&Norm&No\\
46 &OU-ISIR - (Speed)(2014)\cite{Mansur2014Gait}&333&333&GR&Sil&Norm&Yes\\
47 &TUM GAID(2014)\cite{Hofmann2014TUM}&305&3050&GR&Audio,RGB, 3D-S,&Norm&Yes\\
48 &MARS(2016)\cite{Zheng2016MARS}&1261&1261&GR&RGB&Norm&No\\
49 &UPCV Gait K2(2016)\cite{Kastaniotis2016Posebased}&20&100&GR&3D-S&Norm&No\\
50 &LPW(2017)\cite{Song2017Regionbased}&2731&2731&GR&RGB&Norm&No\\
51 &OU-ISIR - (Age)(2017)\cite{Xu2017OUISIR}&63846&N/S&GR&GEI, Sil&Norm&No\\
52 &Aerial Gait Dataset(2018)\cite{Perera2018Human}&2&17&GR&RGB&Norm&Yes\\
53 &iLIDS-VID(2018)\cite{Li2018Unsupervised}&300&600&GR&RGB&Norm&Yes\\
54 &OU-ISIR - (Bag)(2018)\cite{Uddin2018OUISIR}&62528&N/S&GR&Sil&Norm&No\\
55 &OU-ISIR MVLP(2018)\cite{Takemura2018Multiview}&10307&144298&GR&GEI, Sil&Norm&No\\
56 &FVG(2019)\cite{Zhang2019Gait}&226&2856&GR&RGB&Norm&Yes\\
57 &GPJATK(2019)\cite{Kwolek2019Calibrated}&32&664&GR&3D Mocap, RGB&Norm&Yes\\
58 &GRIDDS(2019)\cite{Nunes2019GRIDDS}&35&350&GR&3D-S&Norm&Yes\\
59 &KinectREID(2019)\cite{Gianaria2019Robust}&71&&GR&3D-S&Norm&No\\
60 &KinectUNITO’13 (2019)\cite{Gianaria2019Robust}&20 &400&GR&3D-S&Norm&No\\
61 &KinectUNITO’16(2019)\cite{Gianaria2019Robust}&10&160&GR&3D-S&Norm&No\\
62 &OutdoorGait(2019)\cite{Song2019GaitNet}&138&4968&GR&RGB&Norm&Yes\\
63 &SMVDU-Multi-Gait(2019)\cite{Singh2019MultiGait}&20&240&GR&RGB&Norm&Yes\\
64 &OUMVLP-Pose(2020)\cite{An2020Performance}&10307&144298&GR&2D-S&Norm&No\\
65 &Buaa-Duke-Gait(2021)\cite{Zhang2023RealGait}&1404&2319&GR&GEI, RGB, Sil&Norm&No\\
66 &GREW(2021)\cite{Zhu2022Gait}&26345&128671&GR&2D-S, 3D-S, GEI, OF, Sil&Norm&No\\
67 &ReSGait(2021)\cite{Mu2021ReSGait}&172&870&GR&Sil, Skeleton&Norm&No\\
68 &SACV-Gait(2021)\cite{Masood2021Appearance}&121&121&GR&RGB&Norm&Yes\\
69 &BRIAR(2022)\cite{CornettIII2022Expanding}&1,000(approx.)&N/S&GR&RGB&Norm&Yes\\
71 &CASIA - E(2022)\cite{Song2023CASIAE}&1014&778752&GR&RGB&Norm&Yes\\
72 &CU Denver Gait(2022)\cite{Konz2022STDeepGait}&100& 3,087&GR&3D-S&Norm&No\\
73 &Depression and Emotion Gait(2022)\cite{Yang2022Data}&95&380&GR&3D-S&Norm&No\\
74 &Gait3D(2022)\cite{Zheng2022Gait}&4000&25000&GR&3D Mesh, 3D-S, Sil&Norm&No\\
75 &OUMVLP-Mesh(2022)\cite{Li2022MultiView}&10307&144298&GR&3D Mesh&Norm&No\\
76 &ST-DeepGait(2022)\cite{Konz2022STDeepGait}&100&3087&GR&3D-S&Norm&No\\
77 &TTG-200(2022)\cite{Liang2022GaitEdge}&200&14198&GR&Sil&Norm&No\\
78 &UAV-Gait(2022)\cite{Ding2022Dataset}&202&9, 898&GR&3D-S, RGB, Sil&Norm&No\\
79 &OG RGB+D(2023)\cite{Li2023multimodala}&96&58752&GR&Depth, RGB, Sil&Norm&No\\
80 &VersatileGait(2023)\cite{Zhang2023LargeScale}&10000&1 mil (approx.)&GR&Sil&Norm&No\\
81 &CASIA - B(2006)\cite{Yu2006framework}&124&13640&GR &RGB, Sil&Norm&Yes\\
82 &Zago(2020)\cite{Zago20203D}&2&N/S&MMC&3D Mocap, RGB&Norm&No\\
83 &Needham(2021)\cite{Needham2021accuracy}&15&150&MMC&3D Mocap, RGB&Norm&No\\
84 &FusionGait(2022)\cite{Yamamoto2022Verification}&16&16&MMC&IMU, RGB&Norm&No\\
85 &Liang(2022)\cite{Liang2022reliability}&30&30&MMC &2D-S, 3D Mocap, RGB&Norm&No\\
86 &Fleuret(2008)\cite{Fleuret2008Multicamera}&N/S&N/S&MOT&RGB&Norm&Yes\\
87 &PETS 2009(2009)\cite{Ferryman2009PETS2009}&N/S&N/S&MOT&RGB&Norm&Yes\\
88 &USC Campus(2010)\cite{Kuo2010Intercamera}&N/S&N/S&MOT&RGB&Norm&Yes\\
89 &Berclaz(2011)\cite{Berclaz2011Multiple}&N/S&N/S&MOT&RGB&Norm&Yes\\
90 &2DMOT2015(2015)\cite{Leal-Taixe2015MOTChallenge}&N/S&22&MOT&RGB&Norm&Yes\\
91 &DukeMTMC(2016)\cite{Ristani2016Performance}&N/S&N/S&MOT&RGB&Norm&Yes\\
92 &MOT16(2016)\cite{Milan2016MOT16}&N/S&14&MOT&RGB&Norm&Yes\\
93 &MOT20(2020)\cite{Dendorfer2020MOT20}&N/S&8&MOT&RGB&Norm&Yes\\
94 &Pedestrian Direction(2017)\cite{Dominguez-Sanchez2017Pedestrian}&N/S&N/S&Pedestrian Tracking&RGB&Norm&Yes\\
95 &PCG dataset(2020)\cite{Yamada2020Gaitbased}&30&30&Pedestrian Tracking&LiDAR&Norm&No\\
96 &PTB-TIR(2020)\cite{Liu2020PTBTIR}&N/S&60&Pedestrian Tracking&TIR&Norm&No\\
\bottomrule
\multicolumn{8}{p{850pt}}
{
\textbf{NOTE}: 2D-S - 2D Skeleton ;3D-S - 3D Skeleton;N/S - Not Stated; AR - Action Recognition; AP - Abnormality or Pathology Detection; Abnorm - Abnormal; AQA - Action Quality Assessment; BA - Biomechanical Analysis; GR - Gait Recognition; MMC - Markerless Motion Capture; Multi-Object-Tracking - MOT; Norm - Normal; Path - Pathological.
}
\end{tabular}}
\vspace{-10pt}
\end{table*}

Figure \ref{fig:number_datasets} summarises various computer vision tasks related to gait actions. Gait recognition has been the primary focus of many of the current gait-related datasets \cite{Sethi2022comprehensive}. These datasets typically consist of gait sequences presented in silhouettes format derived from RGB video data. However, despite using RGB methods such as closed-circuit television (CCTV) or video to capture initial gait sequences, many of these datasets do not provide access to the original RGB data due to issues of data storage and privacy. 

Since the primary task is biometric identification based on gait patterns, data collection often assumes representative general populations displaying typical variations of normal gait patterns. Consequently, these datasets lack representation of abnormal or pathological gait types and lack annotations that would facilitate clinical gait analysis. However, they do provide ample examples of normative gait patterns. For example \cite{Li2022MultiView} provides a dataset comprising over 10,000 subjects captured in multiple views, represented with 2D skeletons and 3D body mesh.

There exists a notable number of gait datasets created for the task of abnormality/pathology detection. However many of these datasets suffer from several limitations, including a small number of subjects and sequences, limited variety of abnormal gaits, restricted representation of pathologies, and scarcity of RGB data availability. Privacy concerns further exacerbate these limitations, as many datasets are either unavailable or only accessible in reduced dimensional data types such as key point coordinates. As a result, the development of using rich features extracted from RGB data is hindered.

Biomechanical analysis, which involves analysis of individual parts of the human body to understand human movement in sporting and everyday activities, shares similarities with abnormality/pathology detection. However, its primary aim is to measure and describe human movement. While there are a few papers providing gait data for biomechanical analysis, similar privacy concerns limit the availability of these datasets or restrict them to reduced dimensional data types. 

Action recognition, a well-known task in computer vision, classifies human actions in images and video. Recent developments in human action recognition have been driven by large datasets. However despite the large number of action classes, many either lack a dedicated walking class or have a walking class that is not suitable for CGA due to inconsistent gait cycles or noisy data. 
Multi-object tracking, a task involving the identification of people and objects from single or multiple camera views, shares a use case with gait recognition in terms of surveillance and person identification. Similarly, pedestrian tracking, used in robotics and autonomous vehicles, tracks the path of a person walking. 
Many available multi-object tracking and pedestrian datasets provide bounding box annotations for people walking within a scene. This data has an advantage as subjects are typically observed in natural environments, performing natural gait. However, the absence of clinical annotations and the presence of objects such as bags carried by subjects can impact gait patterns.

Current gait datasets exhibit variability in several aspects, including action sequence lengths, the presence of subjects without pathological gait, variations in camera viewing angles, and a range of covariate factors such as gait speeds, clothing items, and carrying of objects. 
Table \ref{table:Normal-datasets} offers a comprehensive summary of 96 video datasets with normal gait actions. Meanwhile, Table \ref{tab:Abnormal-datasets} provides a detailed overview of 61 gait datasets containing abnormal or pathological gait patterns, highlighting the various types of abnormalities and pathologies represented in the datasets.

\begin{table*}[!t]
\vspace{-5pt}
\centering
\caption{Datasets with Abnormal Gait Actions and Activities}
\label{tab:Abnormal-datasets}
\resizebox{0.98\textwidth}{!}{%
\begin{tabular}{l|l c c c l l c}
\toprule
 & \textbf{Dataset (year)}  & \textbf{Total Subjects} & \textbf{Total Gait Sequences} & \textbf{Original Purpose} & \textbf{Data Modalities} & \textbf{Gait Type} & \textbf{RGB Available}\\ 
\midrule
1 &OMally(1997)\cite{OMalley1997Fuzzy}&156&N/S&AD&3D Mocap&Norm/Path (CP-child)&No\\
2 &Bauckhage(2005)\cite{Bauckhage2005Detecting}&7&N/S&AD&RGB, Sil&Norm/Abnorm(Msk)&No\\
3 &Wolf(2006)\cite{Wolf2006Automated}&64&N/S&AD&3D Mocap&Norm/Path (CP-child)&No\\
4 &DAI2(2015)\cite{Nieto-Hidalgo2017Classification}&5&75&AD&GEI, RGB, Sil&Norm/Abnorm (Msk)&No\\
5 &DGD(2015)\cite{Chaaraoui2015Abnormal}&7&56&AD&3D-S&Norm/Abnorm(Msk)&No\\
6 &Prochazka2015(2015)\cite{Prochazka2015Bayesian}&51&N/S&AD&3D-S&Norm /Norm Variant ( Elderly)/Path (PD)&No\\
7 &SPHERE2015(2016)\cite{Tao2016comparative}&10&40&AD&3D-S&Norm/Path (PD)/Path (stroke)&No\\
8 &Diplegia(2017)\cite{Bergamini2017Signal}&156&N/S&AD&3D Mocap&Norm/Path (CP)&No\\
9 &Gholami(2017)\cite{Gholami2017Microsoft}&20&20&AD&3D-S&Norm/Path(MS)&No\\
10 &Laet(2017)\cite{Laet2017Does}&356&1719&AD&3D Mocap&Norm/Path (CP-child)&No\\
11 &Bei(2018)\cite{Bei2018Movement}&120&N/S&AD&3D-S&Norm/Abnorm(Msk)&No\\
12 &INIT Gait(2018)\cite{Ortells2018Visionbased}&10&160&AD&Sil&Norm/Abnorm (Msk)&No\\
13 &Kozolow(2018)\cite{Kozlow2018Gait}&28&168&AD&3D-S&Norm/Abnorm(Msk)&No\\
14 &Li(2018)\cite{Li2018Classification}&42&N/S&AD&3D-S, RGB&Norm/Path (CP)/Path (PD)&No\\
15 &Walking Gait(2018)\cite{Nguyen2018Walking}&9&81&AD&3D-S, PC, Sil&Norm/Abnorm (Msk)&No\\
16 &Amini(2019)\cite{Amini2019Using}&45&N/S&AD&3D-S&Norm/Path(PD)&No\\
17 &Fang(2019)\cite{Fang2019Depression}&3669&N/S&AD&3D-S&Norm/Path(MHD)&No\\
18 &Lee(2019)\cite{Lee2019Abnormal}&N/S&N/S&AD&3D-S&Norm/Abnorm(Msk)&No\\
19 &MMGS(2019)\cite{Khokhlova2019Normal}&27&489&AD&3D-S, Depth, Sil&Norm/Abnorm (Msk)&No\\
20 &SMAD(2019)\cite{Sardari2019ViewInvariant}&19&80&AD&RGB&Norm/Path (PD)/Path (Msk)/Path (Stroke)&Yes\\
21 &Caicedo(2020)\cite{Caicedo2020Dataset}&44&N/S&AD&3D Mocap&Norm/Norm Variant (Elderly)&No\\
22 &Chakraborty(2020)\cite{Chakraborty2020Gait}&30&N/S&AD&3D-S&Norm/Path (CP-child)&No\\
23 &DeOliveiraSilva(2020)\cite{DeOliveiraSilva2020Gait}&63&N/S&AD&RGB&Norm Variant ( Elderly)/Path (CI)/Path (PD)&No\\
24 &EJMQA(2020)\cite{Elkholy2020Efficient}&43&N/S&AD&3D-S&Norm/Abnorm(Msk)&No\\
25 &GAIT-IST(2020)\cite{Loureiro2020Using}&10&360&AD&2D-S, GEI, RGB, SEI, Sil&Norm/Path(CP)/Path(NP)/Path(PD)&No\\
26 &Kondragunta(2020)\cite{Kondragunta2020Gait}&20&N/S&AD&3D-S&Norm/Path (CI)&No\\
27 &NeuroSynGait(2020)\cite{Goyal2020Detection}&4&258&AD&RGB&Norm/Path (PD)/Path (CP)/Path (HC)&No\\
28 &Pachon-Suescun(2020)\cite{Pachon-Suescun2020Abnormal}&N/S&220&AD&3D-S&Norm/Path(CP)/Path(PD)&No\\
29 &Path Gait(2020)\cite{Jun2020Pathological}&10&720&AD&3D-S&Norm/Path (Msk)&No\\
30 &Seifallahi(2020)\cite{Seifallahi2020Alzheimer}&60&60&AD&3D-S&Norm/Path (AD)&No\\
31 &PD dataset(2020)\cite{Dadashzadeh2020Exploring}&25&1058&AD&MB, OF, RGB &Path (PD)&No\\
32 &Tsukagoshi(2020)\cite{Tsukagoshi2020Noninvasive}&75&N/S&AD&3D-S&Norm/Path(ND)/Path(PD)&No\\
33 &Gait and Full Body Movement(2021)\cite{Al-Jubouri2021Gait}&59&109&AD&3D-S, RGB &Norm/Path (Autism)&No\\
34 &GAIT-IT(2021)\cite{Albuquerque2021Spatiotemporal}&21&828&AD&2D-S, GEI, SEI, Sil&Norm/Path(CP)/Path(NP)/Path(PD)&No\\
35 &Wang(2021)\cite{Wang2021Gait}&98&N/S&AD&3D-S&Norm/Path(MHD)&No\\
36 &Beside Gait(2023)\cite{Cheriet2023Multispeed}&43&115&AD&2D-S&Norm/Path(ND)&No\\
37 &Simulated Path Gait Dataset(2023)\cite{Jun2023Hybrid}&12&1440&AD&3D-S&Norm/Path (Msk)&No\\
38 &Vestibular Disorder Gait(2023)\cite{Jun2023Hybrid}&132&482&AD&3D-S&Norm/Path (VD)&No\\
39 &Wang(2024)\cite{Wang2024VideoBased}&44&92&AD&RGB&Norm/Path (NC)/Path (PD)&No\\
40 &Zhou(2024)\cite{Zhou2024Portable}&14&224&AD&RGB&Norm/Path(Stroke)&No\\
41 &QMAR(2020)\cite{Sardari2020VINet}&38&912&AQA&RGB, 3D-S&Norm/Path (PD)/Path (Stroke)&Yes\\
42 &UoY Emotion(2014)\cite{Keefe2014database}&29&N/S&AR&RGB&Norm/Norm Variant (Emotions)&Yes\\
43 &Mostayed(2008)\cite{Mostayed2008Abnormal}&79&N/S&BA&3D Mocap&Norm/Abnorm(Msk)&No\\
44 &Qu(2011)\cite{Qu2011Effects}&12&N/S&BA&3D Mocap&Norm/Norm Variant (Carrying Object)&No\\
45 &Moore(2015)\cite{Moore2015elaborate}&15&45&BA&3D Mocap&Norm/Norm Variant (Purturbed)&No\\
46 &Nguyen(2016)\cite{Nguyen2016SkeletonBased}&5&N/S&BA&3D-S&Norm/Abnorm(Msk)&No\\
47 &Kin-FOG(2019)\cite{Soltaninejad2019KinFOG}&5&N/S&BA&3D-S&Norm/Path(PD)&No\\
48 &Latorre(2019)\cite{Latorre2019Gait}&437&N/S&BA&3D-S&Norm/Path(Stroke)&No\\
49 &TTR-FAP Gait(2019)\cite{Vilas-Boas2019Validation}&10&10&BA&3D-S, RGB, Sil&Norm/Path(ND)&No\\
50 &Rana(2021)\cite{Rana20213D}&24&N/S&BA&3D-S,  IR-UWB&Norm/Abnorm(Msk)&No\\
51 &Bertaux(2022)\cite{Bertaux2022Gait}&186&186&BA&3D Mocap&Norm/Path (Msk)&No\\
52 &Lonini(2022)\cite{Lonini2022VideoBaseda}&8&N/S&BA&RGB&Path(Stroke)&Yes\\
53 &M2OCEAN(2023)\cite{VanCriekinge2023fullbody}&138&N/S&BA&3D Mocap, EMG&Path(Stroke)&No\\
54 &EffortShape(2012)\cite{Gross2012EffortShape}&16&142&GR&3D Mocap, RGB&Norm/Norm Variant (Emotions)&Yes\\
55 &Venture Gait(2014)\cite{Venture2014Recognizing}&4&100&GR&3D Mocap, RGB&Norm/Norm Variant (Emotions)&Yes\\
56 &E-Gait(2020)\cite{Bhattacharya2020STEP}&90&3177&GR&3D Mocap, RGB&Norm/Norm Variant (Emotions)&Yes\\
57 &EMOGAIT(2021)\cite{sheng2021multi}&60&1440&GR&RGB&Norm/Norm Variant (Emotions)&Yes\\
58 &EmoGait3d(2021)\cite{Yu2021Gaitbased}&27&742&GR&3D-S&Norm/Norm Variant (Emotions)&No\\
59 &ENSAM(2021)\cite{Vafadar2021novel}&31&31&MMC&3D Mocap, RGB&Norm/Norm Variant (Children)/Path (Msk)&No\\
60 &ENSAM-Extended(2022)\cite{Vafadar2022Assessment}&41&41&MMC&3D Mocap, RGB&Norm/Norm Variant (Children)/Path (Msk)&No\\
61 &TOAGA(2022)\cite{Mehdizadeh2022Toronto}&14&14&MMC&3D Mocap, RGB&Norm/Norm Variant (Elderly)&No\\
\bottomrule
\multicolumn{8}{p{900pt}}
{
\textbf{NOTE}: Quality Assessment; AR - Action Recognition; BA - Biomechanical Analysis; GR - Gait Recognition; MMC - Markerless Motion Capture; N/S - Not Stated; Norm - Normal; Norm Variant ( Elderly) - Normal Varient - Elderly Person; Norm Variant (Carrying Object) - Normal Varient - Person Carrying Object; Norm Variant (Children) - Normal Varient - Person is Child; Norm Variant (Emotions) - Normal Varient - Person With Emotion; Norm Variant (Purturbed) - Normal Varient - Person With External Forces; Path - Pathological; Path (AD) - Alzheimer's Disorder Pathology; Path (Autism) - Autism Spectrum Disorder Pathology; Path (CI) - Cognitive Impairment Pathology ; Path (CP) - Cerebral Palsy Pathology; Path (CP-child) - Cerebral Palsy In Child Pathology; Path (HC) - Huntington's Chorea Pathology; Path (Msk) - Musculoskeletal Pathology; Path (NC) - Neurocognitive Disorder Pathologies; Path (PD) - Parkinson's Disease Pathology; Path (VD) - Vestibular Disorder Pathology; Path(MHD) - Mental Health Disorder Pathology; Path(MS) - Multiple Sclerosis Pathology; Path(ND) - Neurodegenrative Pathology; Path(NP) - Neuropathic Gait Pathology; Path(Stroke) - Stroke Pathology; 
}
\end{tabular}}
\vspace{-4pt}
\end{table*}


Datasets with abnormal or pathological gait types are typically less diverse with smaller number of subjects and sequence size. Often they are not publicly available due to medical regulations. This is particularly the case for RGB data. Often there are challenges in acquiring gait data from populations who demonstrate abnormal gait. Other limiting factors include stringent ethical and regulatory processes as well as issues related to privacy. Collecting data from people with pathologies, typically from a medical environment, is challenging due to issues of accessibility and privacy. There can be barriers for people with mobility issues, common in gait abnormalities, to travel and access institutions such as motion capture labs. As a result, many existing gait datasets with abnormal gaits have been produced using actors to simulate or wear equipment to simulate an abnormal gait. In many cases, authors have collected gait data specific to their objective and have not made this public or have provided reduced dimensional data, such as only providing key point data instead of full RGB images \cite{Debnath2022review}. RGB data is important for current computer vision research, the omission of this data from publication is potentially limiting development in healthcare and related fields of human movement analysis such as CGA. 


\subsection{GAVD: Gait Abnormality in Video Dataset}

In response to the demand for a large and representative video dataset featuring abnormal gait patterns, we present, as the main contribution of this paper, the Gait Abnormality in Video Dataset (GAVD). GAVD stands as the largest collection of online-accessible abnormal gait videos annotated by  expert clinicians for clinical analysis (See Table \ref{tab:GAVD Dataset Summary}). 



\begin{table}[!t] 
\centering
\caption{GAVD Dataset Features Summary}
\label{tab:GAVD Dataset Summary}
\resizebox{0.6\textwidth}{!}{%
\begin{tabular}{l c c c}
\toprule
\textbf{Feature} & \textbf{Normal} & \textbf{Abnormal} & \textbf{Total} \\
\midrule
URL Links       & 41    & 411   & 452 \\
Gait Sequences  & 291   & 1,583 & 1,874 \\
Frames          & 41,340& 416,776&458,116 \\
\bottomrule
\end{tabular}
}
\end{table}

\subsubsection{Data collection and annotation}
Publicly available online videos of normal and abnormal gaits were manually sourced from public domain websites and video portals such as Youtube. Expert clinicians screened and annotated relevant videos to ensure clinically accurate identification of abnormal gaits. Studies like \cite{Toro2003review} and \cite{Lee2020Perceiving} have highlighted the consistent ability of expert clinicians, such as physiotherapists, to accurately identify abnormal gait events. 

Abnormal gaits are defined as gait patterns that significantly deviate from statistical or biomechanical norms based on subjective appearance. These patterns consist of gaits resulting from pathologies, simulated abnormalities, and exercises such as toe walking which consists of walking with modified posture. 
It is important to note that the term ``abnormal'' does not imply negativity or inferiority but rather indicates deviation from typical gait patterns. Subjective identification of abnormal gait is often the initial step in diagnosing issues and pathologies\cite{baker2006gait}. 

Clinical experts screen and identify normal and abnormal gait sequences from over 450 online videos. Each video is further annotated frame-by-frame focusing on subjects demonstrating gait sequences consisting of at least two gait cycles with minimal visual obstruction, walking in a singular direction relative to the camera. Bounding box coordinates were annotated for each subject in each frame, with manual annotation of the starting and ending frames of each sequence to reduce the impact of noisy and irrelevant data inherent in RGB video captured in uncontrolled environments. In this way, irrelevant parts of videos can be ignored, helping focus on uninterrupted gait sequences useful for the task of clinical gait analysis. 
Additionally, each sequence was labelled with clinically relevant gait classes, including normal and abnormal, with further sub-classifications of abnormal sequences into idiopathic and pathological presentations (See Table \ref{tab:GAVD_gait_classes}).  

\subsubsection{GAVD dataset statistics}
The GAVD Dataset provides the URL links to over 450 videos, each with corresponding clinical annotations of normal and abnormal gait sequences. The dataset comprises 416,776 frames exhibiting abnormal gait patterns, with 458,116 clinically annotated bounding box coordinates provided alongside camera view annotations.

\begin{figure}
    \centering
    \includegraphics[width=1\linewidth]{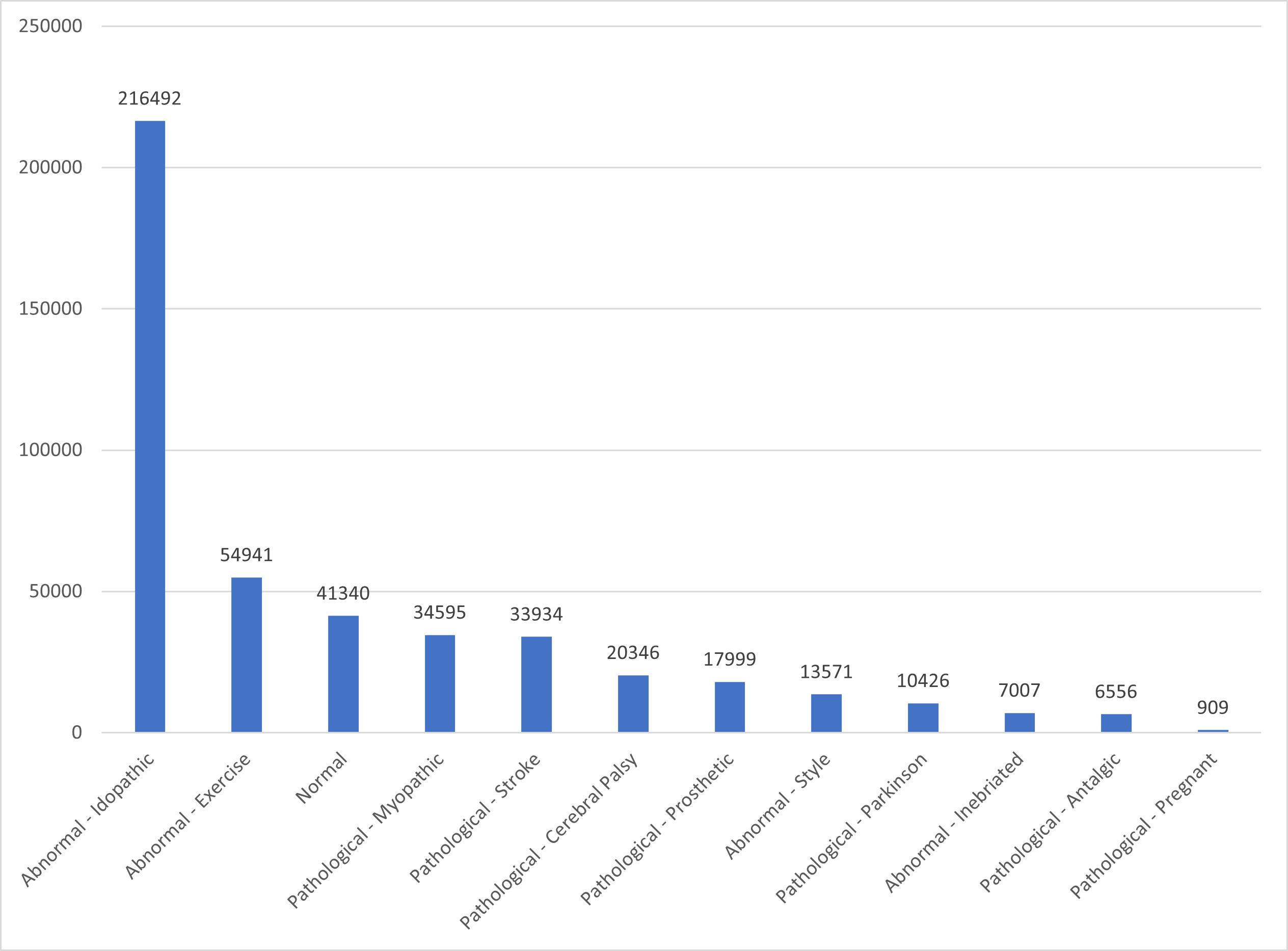}
    \caption{Distribution of Frames per Subclass in GAVD}
    \label{fig:subclass_gavd}
\end{figure}

GAVD offers a diverse and representative mix of covariates including age, height, gender, ethnicity, body shape, clothing, and indoor and outdoor environments. Additionally, the dataset provides 11 sub-classifications of abnormal gait patterns for future development in CGA. The distribution of abnormal gait sub-classes is depicted in Figure \ref{fig:subclass_gavd}, with the majority labelled as ``idiopathic'' due to the unclear cause of the identified gait abnormality.


\begin{table*} 
\caption{GAVD Gait Class Definitions}
\label{tab:GAVD_gait_classes}
\centering
\resizebox{0.99\textwidth}{!}{%
\begin{tabular}{
l l >{\raggedright\arraybackslash}p{4.5cm} >{\raggedright\arraybackslash}p{3cm} >{\centering\arraybackslash}p{2.5cm}
} 
\toprule
\textbf{Class}&\textbf{Sub-Class}& \textbf{Definitions}& \textbf{Example}&\textbf{No. of Frames} \\ 
\midrule
Abnormal & Exercise& Abnormal gait presentation due to deliberate exercise.& Person walking on heels of leg.&54,941\\ 
\midrule
Abnormal & Idiopathic& Abnormal gait presentation without clear cause.& Person walking without bending knee.&216,492\\ 
\midrule 
Abnormal & Inebriated& Person walking with the appearance of being under influence of a substance.& Person unable to walking in straight line. &7,007\\ 
\midrule 
Abnormal & Style& Person walking with exaggerated emotional or caricature styles.& Person walking "sad" with slumped posture. &13,571\\ 
\midrule 
Normal& Normal &Person walking without clinically obvious presentation of abnormality.& Person walking in the street with natural appearance in gait.&41,340\\ 
\midrule 
Pathological & Antalgic& Person walking with clinically typical signs of pain during gait.& Person hobbles and expresses pain during walking.&6,556\\ 
\midrule 
Pathological & Cerebral Palsy& Person walking with clinically typical signs of single limb or bilateral dystonia or ataxia in gait.& Person walking with dystonia.&20,346\\ 
\midrule 
Pathological & Myopathic& Person walking with clinically typical signs of myopathy or lower limb muscle weakness in gait.& Person walking and hitching hip.&34,595\\ 
\midrule
Pathological & Parkinson& Person walking with clinically typical signs of shuffling or festinating gait.& Person shuffles to walk.&10,426\\
\midrule
Pathological & Pregnant& Person walking with clinically typical signs of pregnancy during gait.& Person walks with pregnant appearance, wide base support.&909\\
\midrule
Pathological & Prosthetic& Person walking with clinically typical signs of single limb or bilateral prosthetic in gait.& Person walking with below-knee prosthetic.&17,999\\
\midrule
Pathological & Stroke& Person walking with clinically typical signs of post stroke contracture in gait.& Person walks with contracture.&33,934\\
\bottomrule
\end{tabular}}
\end{table*}

\begin{figure}[h]
    \centering
     \includegraphics[width=0.7\linewidth]{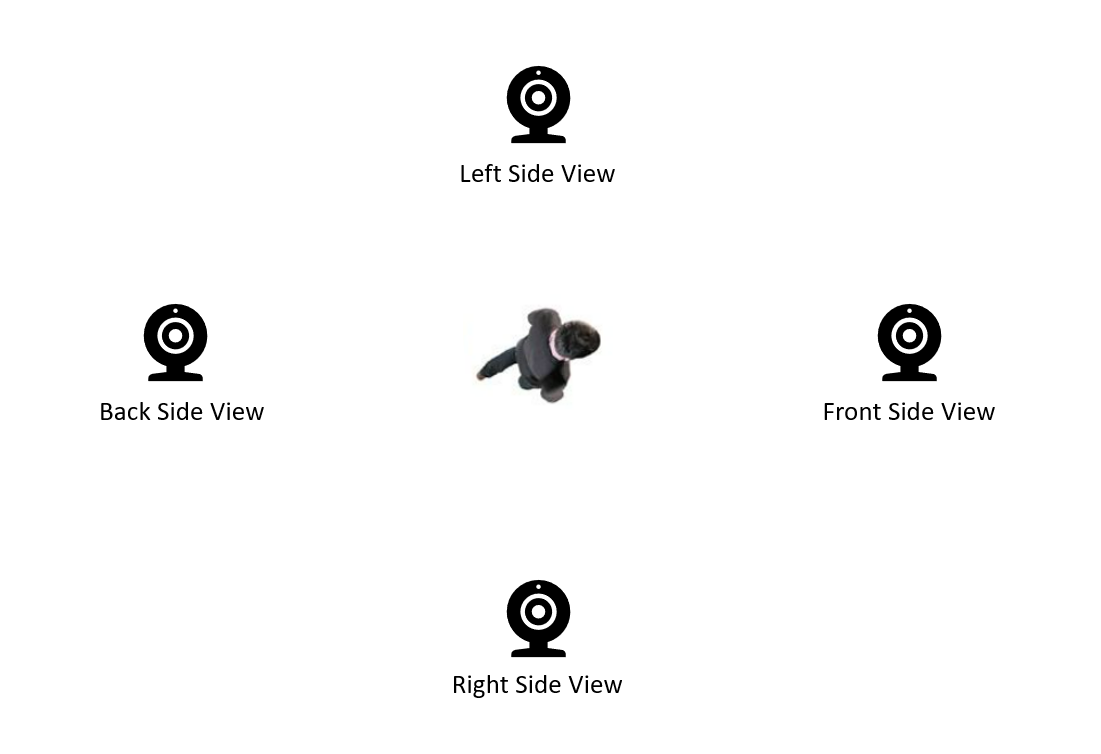}
\caption{Person-centric Camera View}
    \label{fig:Person-centric Camera View}
\end{figure}

Furthermore, person-centric camera view labels are provided to aid the development of view-invariant identification and classification tasks  (See Figure \ref{fig:Person-centric Camera View}). GAVD ensures a balanced distribution of sequences per person-centric camera view as illustrated in Table \ref{tab:person-centric camera}.

\begin{table}
\centering
\caption{GAVD Dataset: Distribution of Sequences per Person-Centric View}
\resizebox{0.6\textwidth}{!}{%
\begin{tabular}{c|c} 
\toprule
\textbf{Person-Centric View} &  \textbf{Number of sequences}\\ \hline 
Front view&  522\\ \hline 
Back view&  319\\ \hline 
Left side view&537\\ \hline 
Right side view&499\\ \hline
Total &1877\\
\bottomrule
\end{tabular}}
\label{tab:person-centric camera}
\end{table}

\subsection{Clinical Abnormality Simulated Dataset}

To validate the GAVD dataset, we conducted a test using data collected in a controlled environment known as the Clinical Abnormality Simulated Dataset (CASD).
CASD simulates clinical conditions under the supervision of a clinician, allowing us to extend computer vision gait analysis research towards clinically relevant scenarios.
While integrating computer vision into healthcare contexts can pose various challenges, accessing simulated clinical data helps bridge this gap. 
However, we have not been authorised for public release of this data at this point. 
In the CASD, 20 participants performed four varied gait conditions: 
i) Walking Forwards (Normal); 
ii) Walking Backwards (Abnormal); 
iii) Walking Forwards Vision impaired (Abnormal); and 
iv) Walking Backwards Vision impaired (Abnormal). 

To induce vision impairment, participants wore a fitted blindfold garment effectively limiting their visual senses. This method, widely documented in research in research \cite{Iosa2012Effects}, reliably induces subtle changes in gait patterns non-invasively. Additionally, walking backwards creates significant alterations in gait kinematics \cite{Lee2013Kinematic} and muscle activation patterns \cite{Abdelraouf2019Backward} that deviate from typical forward gait. 

Each participant performed 8 to 10 laps for each condition on a five-meter walkway placed indoors, featuring a flat surface and consistent lighting conditions. Two cameras were placed on the simulated clinical gait walkway, perpendicular to each other. 
Camera 1, placed perpendicular to the walking path, captured left and right side views, while Camera 2, aligned with the walking path, recorded front and back side views. 
Synchronisation between the two cameras ensured simultaneous capture of the right side with backside and the left side with front side views.

\section{Experimental studies}

An experimental study was performed to validate the efficacy of our dataset, GAVD, for CGA tasks in computer vision and to identify potential areas for future research. 
Given the promising potential of region-based action recognition models in identifying gait abnormalities from video data, our focus was on utilising such models for CGA. 
To this end, we employed two region-based human action recognition models:  SlowFast\cite{Feichtenhofer2019SlowFast} and Temporal Segment Network (TSN)\cite{Wang2016Temporal}.
These models were fine-tuned for the binary classification task of recognising abnormal gaits from normal gaits, which is crucial for the initial identification of gait issues and possible pathologies. 
SlowFast and TSN were selected for their efficiency and effectiveness in learning from entire action video clips, such as those collected in GAVD.

\subsection{Fine-tuning action recognition with GAVD }
For fine-tuning action recognition models, we randomly selected 582 sequences (291 Normal and 291 Abnormal) from 1874 gait sequences in the GAVD dataset. These selected sequences served as the training and validation data for our models.

To fine-tune the models, we utilised pre-trained models from the Kinetics-400 dataset \cite{Kay2017Kinetics} and their corresponding weights and configurations available in the mmaction2\cite{2020mmaction2} open-source toolbox for video understanding. 
Due to the inherent noise in videos from uncontrolled environments, we pre-processed the videos to focus attention solely on the actions of the individual in the frame. 
This involved annotating bounding box around the subject of interest and applying padding to mask the remaining frame.

We employed a stratified 10-fold cross-validation protocol was used to fine-tune the models using 9-folds to train and one fold held out for validation. 
Both TSN and SlowFast models were trained for up to 200 epochs utilising a batch size of 16 and default parameters. Throughout the training process, the models showed consistent accuracy and loss when validated against the validation set. This consistency demonstrated the robustness of the data and minimised bias from outlying data. 

\subsection{Evaluation protocol}
The trained TSN and Slowfast models are evaluated against unseen data in the test set to demonstrate the use of video from the GAVD dataset and region-based action recognition for clinical gait analysis. 

Unseen data from three distinct sources are used to assess the versatility of the models. 
The first unseen data is a subset of GAVD consisting of only abnormal gaits patterns. 
The second unseen dataset, GPJATK gait dataset \cite{Kwolek2019Calibrated} consists of only normal gait patterns. 
The third unseen dataset is CASD which provides a clinically feasible gait dataset with controlled background and camera conditions. 
 
\subsubsection{GAVD test set}
GAVD test set consists of randomly selected gait videos from the GAVD dataset that are not used in model training. This includes 1132 unseen gait sequences from the abnormal gait class. This dataset consists of abnormal gait sequences  (n= 1132) only.

\subsubsection{GPJATK test set}
This test set consists of gait videos from the well-known GPJATK dataset~\cite{Kwolek2019Calibrated}. GPJATK gait dataset has 32 subjects with no observable gait abnormalities. 618 gait sequences from multiple fixed camera views are captured in an indoor environment with consistent conditions. The captured gait sequences use four synchronised cameras to capture gait from multiple camera views which ideally allow for testing the influence of person view on abnormality detection. This dataset provides gait sequences only classified in the normal gait class (n= 618).

\subsubsection{CASD test set}
This test set consists of unseen gait sequence videos from the previous described CASD. This dataset consists of 20 subjects performing 738  gait sequences all captured from side view (left and right side).  This dataset provides sequences for both the normal gait class (n= 94) class and the abnormal class (n= 644) in a controlled environment. Subjects performed normal gait actions as well as simulated abnormal gait actions. 

\begin{table*}[!t]
\centering
\caption{Summary of test sets.}
\label{tab:test_s}
\resizebox{0.8\textwidth}{!}{%
\begin{tabular}{l c cl}
\toprule
\textbf{Test set source
}& \textbf{Description}& \textbf{Normal}&\textbf{Abnormal}\\
\midrule
GAVD test set& ~300 subjects, abnormal gaits&0&1132\\
GPJATK test set 2& 32 subjects, normal gait&618&0\\ 
CASD test set 3&  20 subjects, normal and abnormal gaits&94&644\\ 
\bottomrule
\end{tabular}}
\end{table*}

\subsection{Results and analysis}
Both the pre-trained TSN and SlowFast models demonstrated favourable results in identifying abnormal gait patterns in videos captured from a variety of environments, camera positions and recording conditions. 
When tested on the GAVD dataset, both models achieved high accuracy in classifying abnormal gaits, with accuracy of 0.94 and 0.92 for SlowFast and TSN models, respectively. 

There was reduced accuracy for both models when testing on gait videos from GPJATK test dataset when compared to performance on the abnormal class of GAVD in Test set 1.
The SlowFast model performed better, with an accuracy of 0.76 than TSN which achieved an accuracy of only 0.59 in identifying normal gait patterns in GPJATK dataset.  
However, when camera view is taken into consideration both models are able to accurately identify normal gait when the subject is viewed from the front or back as seen from camera views C2 and C4 but struggled with side views C1 and C3 (See Table~\ref{tab:CamerView}). 
This discrepancy raises concerns about the generalisability of the approach for identifying normal walking in unseen datasets and prompts further analysis to explore the impact of camera views on model performance.

Testing on the CASD dataset revealed that both Slow Fast and TSN models tended to classify gait sequences with a side-view perspective as abnormal. Given that the dataset consisted of only side view perspectives and a majority of abnormal class samples, further investigation is required to understand the impact of camera perspective on abnormality detection.   

\begin{table*}[!t]
\centering
\caption{Comparison of model accuracy in gait normality/abnormality classification on test sets.}
\label{tab:Ablation_results}
\resizebox{0.49\textwidth}{!}{%
\begin{tabular}{l c c}
\toprule
\textbf{Test Set} & \textbf{SlowFast} & \textbf{TSN} \\
\midrule
GAVD test set& 0.94&0.92\\
GPJATK test set 2& 0.76&0.59\\ 
CASD test set 3&  0.87&0.73\\ 
\bottomrule
\end{tabular}}
\end{table*}

\begin{table*}[!t]
\centering
\caption{Impact of camera view on detection}
\label{tab:CamerView}
\resizebox{0.85\textwidth}{!}{%
\begin{tabular}{c l c c}
\toprule
\textbf{Camera View} & \textbf{Description} & \textbf{SlowFast} & \textbf{TSN} \\ 
\midrule
C1 & Predominantly Left/Right Side views & 0.66& 0.30\\
C2 & Predominantly Front/Back Views & 0.91& 0.86\\
C3 & Predominantly Left/Right Side views & 0.56& 0.25\\
C4 & Predominantly Front/Back Views & 0.97&0.95 \\
\addlinespace
\bottomrule
\multicolumn{4}{p{320pt}}
{
Models trained on GAVD demonstrate improved accuracy in identifying abnormal movement from front/back views of a person.
}
\end{tabular}}
\end{table*}

\section{Discussion}

The results obtained from the pre-trained action recognition models using the GAVD dataset indicate accurate identification of abnormal gait patterns, as shown in Table~\ref{tab:Ablation_results}. 
This yields promising results for the utility of GAVD and action recognition in clinical gait analysis. 
However, a notable decrease in model performance was observed, with false-positive results indicating the detection of abnormal gaits in sequences viewed from left or right side of the subject. 
Further analysis revealed that the models performed well when the subject walked towards or away from the camera but struggled abnormal gaits when the subject moved perpendicular to the camera view. This suggests the involvement of other factors, such as movement direction or camera perspective, as significant features impacting abnormality detection. This highlights the need for view-invariant methods to assist with accurate abnormal gait detection.  
One possible factor could be the presence of camera lens distortion, particularly amplified at the edges of the frame when observing subjects walking across. 
Future studies would benefit from in-depth analysis regarding the impact of camera lens distortion, camera view, and the potential effects of corrective measures on action classification accuracy.



\section{Future Areas of Research}
Another future research focus is the role of attention. For region-based models such as TSN and SlowFast detection of background features plays. In our testing, we sought to remove the distractions of background and multiple subjects within the frame to focus on detecting the change in target subjects motions only. Future studies of this should investigate the semantic segmentation of a person to truly remove distracting features from background pixels. Further to this, a major challenge using region-based models is being able to disentangling features such as subject identity from non-identity components.
Sepas-Moghaddam and Etemad~\cite{Sepas-Moghaddam2022Deep} identify in their survey, that learning more discriminative gait representations by disentangling features is one of many challenges faced in deep learning in gait recognition. Similarly, for clinical gait analysis features such as subject self-occlusion, varied camera view-points, subject appearance, subject clothing, body part motion, and environmental lighting present challenges for accurately identifying gait-related features. 

In clinical gait analysis, soft kinematics, which are gait features such as stride length and walking speed, are often taken into consideration. In video analysis, soft kinematics are derived from spatial information such as foot location and temporal information. While some methods of computerised clinical gait analysis require input of soft kinematic measures there may be a possibility that spatial information in the form of key-points positions may be able to provide enough information to differentiate normal and abnormal motions due to ubiquitous nature of the human body structure. In a video of a person walking across the field of view, identifying posture or local joint positions and angles at a given gait phase in the gait cycle provides clinicians with crucial information regarding anatomical normality of the movement. In a computer vision context, we foresee future clinical gait analysis to benefit from temporal and spatial normalisation techniques to improve the comparison of gaits. Approaching gait analysis using generalisable inputs such as posture will also likely improve abnormality sub-class identification. 

Both TSN and SlowFast models were primarily developed for action recognition in video for which actions take place over a few seconds. The models were chosen in this study due to their effective convolutional neural network architectures for action recognition in videos with the ability to learn from limited training samples. These models showed promise in their ability to recognise gait abnormality from a video as a whole. However, given their methods in extracting temporal and spatial information, it was expected that such models would have difficulty with the task of frame-wise classification. In actions such as gait, frame-wise classification requires an understanding of typical postures for a given phase of the action cycle. This is an important aspect of clinical gait analysis as it allows clinicians the ability to determine the cause of the gait deviation\cite{baker2006gait}. The phases of gait transition quickly, for example, part of the loading response which is the change between the heel contacting the ground and the foot landing flat happens within less than 10\% of a gait cycle. Given that the typical gait cycle can be 0.89–1.32 seconds for a person aged 18–49 years \cite{whittle2014gait}, this segment of the gait cycle can occur 0.09-0.13 seconds which is less than one frame in video captured at 30 frames per second. As a result, for action recognition systems to recognise fast movement changes as seen in a gait cycle and to be able to detect and report on frame-wise abnormality it would be expected that incorporating alternative approaches such as Temporal To Still Image Learning (T2SIL) \cite{Herath2017Going}. Here temporal information can help in recognising actions from still images which is the case for frame-wise postures. 


\section{Conclusion}
In this paper, we have provided an overview of the current state of gait-related computer vision research. While gait recognition has received extensive attention, we have highlighted the importance of applying computer vision technology in the health domain for clinical gait analysis. 
Models leveraging spatial–temporal visual information hold promise for assisting in clinical gait analysis, aligning with current trends in healthcare practice for movement analysis. However, further efforts are required to provide clinically important information regarding the spatial and temporal nature of abnormalities detected in movement sequences such as gait. 
To progress the development of clinical gait analysis using computer vision, we have introduced a validated dataset, The Gait Abnormality in Video Dataset (GAVD). This accessible and scalable dataset facilitates the use of spatial–temporal visual information, clinically annotated by human movement experts. 
Finally, we emphasise the need to address challenges to improve abnormality detection in human movement, aiming for camera view invariance, pathology adaptation, and frame-wise classification.
These endeavours will contribute to the enhancement of clinical gait analysis using computer vision and pave the way for impactful applications in healthcare.

\section*{Human Research Ethical Approval}
Procedures for dataset collection and analysis were approved by the CSIRO Health and Medical Human Research Ethics Committee (CHMHREC).

\section*{Declaration of generative AI and AI-assisted technologies in the writing process}
During the preparation of this work the authors used ChatGPT 3.5 in order to format tables and improve readability. After using this tool/service, the authors reviewed and edited the content as needed and take full responsibility for the content of the publication.

\bibliographystyle{splncs04}
\bibliography{refs}

\begin{thebibliography}{100}
\providecommand{\url}[1]{\texttt{#1}}
\providecommand{\urlprefix}{URL }
\providecommand{\doi}[1]{https://doi.org/#1}

\bibitem{Abdelraouf2019Backward}
Abdelraouf, O.R., Abdel-Aziem, A.A., Ahmed, A.A., Nassif, N.S., Matar, A.G.: Backward walking effects on activation pattern of leg muscles in young females with patellofemoral pain syndrome. International Journal of Therapy and Rehabilitation  \textbf{26}(1), ~1--9 (Jan 2019), publisher: Mark Allen Group

\bibitem{Ahmedt-Aristizabal2024Deep}
Ahmedt-Aristizabal, D., Armin, M.A., Hayder, Z., Garcia-Cairasco, Norberto.~Petersson, L., Fookes, Clinton.~Denman, S., McGonigal, A.: Deep learning approaches for seizure video analysis: A review. Epilepsy \& Behavior  \textbf{154} (2024)

\bibitem{Al-Jubouri2021Gait}
Al-Jubouri, A.A., Ali, I.H., Rajihy, Y.: Gait and {Full} {Body} {Movement} {Dataset} of {Autistic} {Children} {Classified} by {Rough} {Set} {Classifier}. Journal of Physics: Conference Series  \textbf{1818}(1),  012201 (Mar 2021)

\bibitem{Albuquerque2021Spatiotemporal}
Albuquerque, P., Verlekar, T.T., Correia, P.L., Soares, L.D.: A {Spatiotemporal} {Deep} {Learning} {Approach} for {Automatic} {Pathological} {Gait} {Classification}. Sensors (Basel, Switzerland)  \textbf{21}(18), ~6202 (Sep 2021)

\bibitem{Amini2019Using}
Amini, A., Banitsas, K.: Using {Kinect} v2 to {Control} a {Laser} {Visual} {Cue} {System} to {Improve} the {Mobility} during {Freezing} of {Gait} in {Parkinson}’s {Disease}. Journal of Healthcare Engineering  \textbf{2019}, ~1--8 (Feb 2019)

\bibitem{An2020Performance}
An, W., Yu, S., Makihara, Y., Wu, X., Xu, C., Yu, Y., Liao, R., Yagi, Y.: Performance {Evaluation} of {Model}-{Based} {Gait} on {Multi}-{View} {Very} {Large} {Population} {Database} {With} {Pose} {Sequences}. IEEE Transactions on Biometrics, Behavior, and Identity Science  \textbf{2}(4),  421--430 (Oct 2020), conference Name: IEEE Transactions on Biometrics, Behavior, and Identity Science

\bibitem{Anggraeni2015Gait}
Anggraeni, N.D., Ferryanto, F., Atmojo, S., Mihradi, S., Dirgantara, T., Mahyuddin, A.: Gait {Parameters} {Determination} by {3D} {Motion} {Analyzer} {System} for {Initial} {Indonesian} {Gait} {Database}. Tech. rep., Bandung Institute of Technology (Apr 2015)

\bibitem{Araujo2014Kinect}
Araujo, R.M., Andersson, V.: Kinect {Gait} {Biometry} {Dataset} - data from 164 individuals walking in front of a {X}-{Box} 360 {Kinect} {Sensor}. Unpublished  (2014)

\bibitem{Baker2016Gait}
Baker, R., Esquenazi, A., Benedetti, M., Desloovere, K.: Gait analysis: clinical facts. European Journal of Physical and Rehabilitation Medicine  \textbf{52}(4),  560--574 (2016)

\bibitem{baker2006gait}
Baker, R.: Gait analysis methods in rehabilitation. Journal of neuroengineering and rehabilitation  \textbf{3},  1--10 (2006)

\bibitem{Bauckhage2005Detecting}
Bauckhage, C., Tsotsos, J., Bunn, F.: Detecting abnormal gait. In: The 2nd {Canadian} {Conference} on {Computer} and {Robot} {Vision} ({CRV}'05). pp. 282--288 (May 2005)

\bibitem{Bei2018Movement}
Bei, S., Zhen, Z., Xing, Z., Taocheng, L., Qin, L.: Movement {Disorder} {Detection} via {Adaptively} {Fused} {Gait} {Analysis} {Based} on {Kinect} {Sensors}. IEEE Sensors Journal  \textbf{18}(17),  7305--7314 (Sep 2018)

\bibitem{benson2022real}
Benson, L.C., R{\"a}is{\"a}nen, A.M., Clermont, C.A., Ferber, R.: Is this the real life, or is this just laboratory? a scoping review of imu-based running gait analysis. Sensors  \textbf{22}(5), ~1722 (2022)

\bibitem{Berclaz2011Multiple}
Berclaz, J., Fleuret, F., Türetken, E., Fua, P.: Multiple {Object} {Tracking} {Using} {K}-{Shortest} {Paths} {Optimization}. IEEE transactions on pattern analysis and machine intelligence  \textbf{33}(9),  1806--1819 (Sep 2011)

\bibitem{Bergamini2017Signal}
Bergamini, L., Calderara, S., Bicocchi, N., Ferrari, A., Vitetta, G.: Signal {Processing} and {Machine} {Learning} for {Diplegia} {Classification}. In: Battiato, S., Farinella, G.M., Leo, M., Gallo, G. (eds.) New {Trends} in {Image} {Analysis} and {Processing} – {ICIAP} 2017. pp. 97--108. Lecture {Notes} in {Computer} {Science}, Springer International Publishing, Cham (2017)

\bibitem{Bertaux2022Gait}
Bertaux, A., Gueugnon, M., Moissenet, F., Orliac, B., Martz, P., Maillefert, J.F., Ornetti, P., Laroche, D.: Gait analysis dataset of healthy volunteers and patients before and 6 months after total hip arthroplasty. Scientific Data  \textbf{9}(1), ~399 (2022)

\bibitem{Bhattacharya2020STEP}
Bhattacharya, U., Mittal, T., Chandra, R., Randhavane, T., Bera, A., Manocha, D.: {STEP}: {Spatial} {Temporal} {Graph} {Convolutional} {Networks} for {Emotion} {Perception} from {Gaits}. Proceedings of the AAAI Conference on Artificial Intelligence  \textbf{34}(02),  1342--1350 (Apr 2020), number: 02

\bibitem{bittner1993prediction}
Bittner, V., Weiner, D.H., Yusuf, S., Rogers, W.J., Mcintyre, K.M., Bangdiwala, S.I., Kronenberg, M.W., Kostis, J.B., Kohn, R.M., Guillotte, M., et~al.: Prediction of mortality and morbidity with a 6-minute walk test in patients with left ventricular dysfunction. Jama  \textbf{270}(14),  1702--1707 (1993)

\bibitem{bogle1996use}
Bogle~Thorbahn, L.D., Newton, R.A.: Use of the berg balance test to predict falls in elderly persons. Physical therapy  \textbf{76}(6),  576--583 (1996)

\bibitem{Borras2012Depth}
Borr{\`a}s, R., Lapedriza, {\`A}., Igual, L.: Depth information in human gait analysis: an experimental study on gender recognition. In: Image Analysis and Recognition: 9th International Conference, ICIAR 2012, Aveiro, Portugal, June 25-27, 2012. Proceedings, Part II 9. pp. 98--105. Springer (2012)

\bibitem{Brach2002Identifying}
Brach, J.S., VanSwearingen, J.M., Newman, A.B., Kriska, A.M.: Identifying {Early} {Decline} of {Physical} {Function} in {Community}-{Dwelling} {Older} {Women}: {Performance}-{Based} and {Self}-{Report} {Measures}. Physical Therapy  \textbf{82}(4),  320--328 (Apr 2002)

\bibitem{Heilbron_2015_CVPR}
Caba~Heilbron, F., Escorcia, V., Ghanem, B., Carlos~Niebles, J.: Activitynet: A large-scale video benchmark for human activity understanding. In: Proceedings of the IEEE Conference on Computer Vision and Pattern Recognition (CVPR). pp. 961--970 (June 2015)

\bibitem{Caicedo2020Dataset}
Caicedo, P.E., Rengifo, C.F., Rodriguez, L.E., Sierra, W.A., Gómez, M.C.: Dataset for gait analysis and assessment of fall risk for older adults. Data in Brief  \textbf{33},  106550 (Dec 2020)

\bibitem{Camarena2023Concise}
Camarena, F., Gonzalez-Mendoza, M., Chang, L., Cuevas-Ascencio, R.J.: A {Concise} {Overview} of the {Vision}-based {Human} {Action} {Recognition} {Field} (Feb 2023)

\bibitem{Chaaraoui2015Abnormal}
Chaaraoui, A.A., Padilla-López, J.R., Flórez-Revuelta, F.: Abnormal gait detection with {RGB}-{D} devices using joint motion history features. In: 2015 11th {IEEE} {International} {Conference} and {Workshops} on {Automatic} {Face} and {Gesture} {Recognition} ({FG}). vol.~07, pp.~1--6 (May 2015)

\bibitem{Chakraborty2020Gait}
Chakraborty, S., Thomas, N., Nandy, A.: Gait {Abnormality} {Detection} in {People} with {Cerebral} {Palsy} {Using} an {Uncertainty}-{Based} {State}-{Space} {Model}. In: Krzhizhanovskaya, V.V., Závodszky, G., Lees, M.H., Dongarra, J.J., Sloot, P.M.A., Brissos, S., Teixeira, J. (eds.) Computational {Science} – {ICCS} 2020. pp. 536--549. Lecture {Notes} in {Computer} {Science}, Springer International Publishing, Cham (2020)

\bibitem{chang2010role}
Chang, F.M., Rhodes, J.T., Flynn, K.M., Carollo, J.J.: The role of gait analysis in treating gait abnormalities in cerebral palsy. Orthopedic Clinics  \textbf{41}(4),  489--506 (2010)

\bibitem{Chao2019GaitSet}
Chao, H., He, Y., Zhang, J., Feng, J.: {GaitSet}: regarding gait as a set for cross-view gait recognition. In: Proceedings of the {Thirty}-{Third} {AAAI} {Conference} on {Artificial} {Intelligence} and {Thirty}-{First} {Innovative} {Applications} of {Artificial} {Intelligence} {Conference} and {Ninth} {AAAI} {Symposium} on {Educational} {Advances} in {Artificial} {Intelligence}. pp. 8126--8133. {AAAI}'19/{IAAI}'19/{EAAI}'19, AAAI Press, Honolulu, Hawaii, USA (Jan 2019)

\bibitem{cheriet2023multi}
Cheriet, M., Dentamaro, V., Hamdan, M., Impedovo, D., Pirlo, G.: Multi-speed transformer network for neurodegenerative disease assessment and activity recognition. Computer Methods and Programs in Biomedicine  \textbf{230},  107344 (2023)

\bibitem{Cheriet2023Multispeed}
Cheriet, M., Dentamaro, V., Hamdan, M., Impedovo, D., Pirlo, G.: Multi-speed transformer network for neurodegenerative disease assessment and activity recognition. Computer Methods and Programs in Biomedicine  \textbf{230},  107344 (Mar 2023)

\bibitem{Collins2002Silhouettebased}
Collins, R., Gross, R., Shi, J.: Silhouette-based human identification from body shape and gait. In: Proceedings of {Fifth} {IEEE} {International} {Conference} on {Automatic} {Face} {Gesture} {Recognition}. pp. 366--371 (May 2002)

\bibitem{2020mmaction2}
Contributors, M.: Openmmlab's next generation video understanding toolbox and benchmark. \url{https://github.com/open-mmlab/mmaction2} (2020)

\bibitem{CornettIII2022Expanding}
Cornett~III, D., Brogan, J., Barber, N., Aykac, D., Baird, S., Burchfield, N., Dukes, C., Duncan, A., Ferrell, R., Goddard, J., Jager, G., Larson, M., Murphy, B., Johnson, C., Shelley, I., Srinivas, N., Stockwell, B., Thompson, L., Yohe, M., Zhang, R., Dolvin, S., Santos-Villalobos, H.J., Bolme, D.S.: Expanding {Accurate} {Person} {Recognition} to {New} {Altitudes} and {Ranges}: {The} {BRIAR} {Dataset} (Nov 2022), arXiv:2211.01917 [cs]

\bibitem{Dadashzadeh2020Exploring}
Dadashzadeh, A., Whone, A., Rolinski, M., Mirmehdi, M.: Exploring {Motion} {Boundaries} in an {End}-to-{End} {Network} for {Vision}-based {Parkinson}'s {Severity} {Assessment} (Dec 2020), arXiv:2012.09890 [cs]

\bibitem{Debnath2022review}
Debnath, B., O’Brien, M., Yamaguchi, M., Behera, A.: A review of computer vision-based approaches for physical rehabilitation and assessment. Multimedia Systems  \textbf{28}(1),  209--239 (Feb 2022)

\bibitem{Dendorfer2020MOT20}
Dendorfer, P., Rezatofighi, H., Milan, A., Shi, J., Cremers, D., Reid, I., Roth, S., Schindler, K., Leal-Taixé, L.: {MOT20}: {A} benchmark for multi object tracking in crowded scenes (Mar 2020), arXiv:2003.09003 [cs]

\bibitem{di2020gait}
Di~Biase, L., Di~Santo, A., Caminiti, M.L., De~Liso, A., Shah, S.A., Ricci, L., Di~Lazzaro, V.: Gait analysis in parkinson’s disease: An overview of the most accurate markers for diagnosis and symptoms monitoring. Sensors  \textbf{20}(12), ~3529 (2020)

\bibitem{Ding2022Dataset}
Ding, T., Zhao, Q., Liu, F., Zhang, H., Peng, P.: A {Dataset} and {Method} for {Gait} {Recognition} with {Unmanned} {Aerial} {Vehicless}. In: 2022 {IEEE} {International} {Conference} on {Multimedia} and {Expo} ({ICME}). pp.~1--6 (Jul 2022), iSSN: 1945-788X

\bibitem{Dominguez-Sanchez2017Pedestrian}
Dominguez-Sanchez, A., Cazorla, M., Orts-Escolano, S.: Pedestrian {Movement} {Direction} {Recognition} {Using} {Convolutional} {Neural} {Networks}. IEEE Transactions on Intelligent Transportation Systems  \textbf{18}(12),  3540--3548 (Dec 2017), conference Name: IEEE Transactions on Intelligent Transportation Systems

\bibitem{Elkholy2020Efficient}
Elkholy, A., Hussein, M.E., Gomaa, W., Damen, D., Saba, E.: Efficient and {Robust} {Skeleton}-{Based} {Quality} {Assessment} and {Abnormality} {Detection} in {Human} {Action} {Performance}. IEEE Journal of Biomedical and Health Informatics  \textbf{24}(1),  280--291 (Jan 2020), conference Name: IEEE Journal of Biomedical and Health Informatics

\bibitem{Fan2020GaitPart}
Fan, C., Peng, Y., Cao, C., Liu, X., Hou, S., Chi, J., Huang, Y., Li, Q., He, Z.: {GaitPart}: {Temporal} {Part}-{Based} {Model} for {Gait} {Recognition}. In: 2020 {IEEE}/{CVF} {Conference} on {Computer} {Vision} and {Pattern} {Recognition} ({CVPR}). pp. 14213--14221. IEEE, Seattle, WA, USA (Jun 2020)

\bibitem{Fang2019Depression}
Fang, J., Wang, T., Li, C., Hu, X., Ngai, E., Seet, B.C., Cheng, J., Guo, Y., Jiang, X.: Depression {Prevalence} in {Postgraduate} {Students} and {Its} {Association} {With} {Gait} {Abnormality}. IEEE Access  \textbf{7},  174425--174437 (2019), conference Name: IEEE Access

\bibitem{Feichtenhofer2019SlowFast}
Feichtenhofer, C., Fan, H., Malik, J., He, K.: {SlowFast} {Networks} for {Video} {Recognition} (Oct 2019), arXiv:1812.03982 [cs]

\bibitem{Ferryman2009PETS2009}
Ferryman, J., Shahrokni, A.: {PETS2009}: {Dataset} and challenge. In: 2009 {Twelfth} {IEEE} {International} {Workshop} on {Performance} {Evaluation} of {Tracking} and {Surveillance}. pp.~1--6 (Dec 2009)

\bibitem{Fleuret2008Multicamera}
Fleuret, F., Berclaz, J., Lengagne, R., Fua, P.: Multicamera people tracking with a probabilistic occupancy map. IEEE transactions on pattern analysis and machine intelligence  \textbf{30}(2),  267--282 (Feb 2008)

\bibitem{Fukuchi2018public}
Fukuchi, C.A., Fukuchi, R.K., Duarte, M.: A public dataset of overground and treadmill walking kinematics and kinetics in healthy individuals. PeerJ  \textbf{6},  e4640 (Jan 2018)

\bibitem{Gammulle2017Two}
Gammulle, H., Denman, S., Sridharan, S., Fookes, C.: Two stream lstm: A deep fusion framework for human action recognition. In: 2017 IEEE Winter Conference on Applications of Computer Vision (WACV). pp. 177--186 (2017). \doi{10.1109/WACV.2017.27}

\bibitem{Gao2022GaitD}
Gao, S., Yun, J., Zhao, Y., Liu, L.: Gait-{D}: {Skeleton}-based gait feature decomposition for gait recognition. IET Computer Vision  \textbf{16}(2),  111--125 (2022)

\bibitem{Gholami2017Microsoft}
Gholami, F., Trojan, D.A., Kövecses, J., Haddad, W.M., Gholami, B.: A {Microsoft} {Kinect}-{Based} {Point}-of-{Care} {Gait} {Assessment} {Framework} for {Multiple} {Sclerosis} {Patients}. IEEE Journal of Biomedical and Health Informatics  \textbf{21}(5),  1376--1385 (Sep 2017), conference Name: IEEE Journal of Biomedical and Health Informatics

\bibitem{Gianaria2019Robust}
Gianaria, E., Grangetto, M.: Robust gait identification using {Kinect} dynamic skeleton data. Multimedia Tools and Applications  \textbf{78}(10),  13925--13948 (May 2019)

\bibitem{Gkalelis2009i3DPost}
Gkalelis, N., Kim, H., Hilton, A., Nikolaidis, N., Pitas, I.: The i3dpost multi-view and 3d human action/interaction database. In: 2009 Conference for Visual Media Production. pp. 159--168 (2009). \doi{10.1109/CVMP.2009.19}

\bibitem{Gorelick2007Actions}
Gorelick, L., Blank, M., Shechtman, E., Irani, M., Basri, R.: Actions as {Space}-{Time} {Shapes}. IEEE Transactions on Pattern Analysis and Machine Intelligence  \textbf{29}(12),  2247--2253 (Dec 2007)

\bibitem{Goyal2020Detection}
Goyal, D., Rao~Jerripothula, K., Mittal, A.: Detection of {Gait} {Abnormalities} caused by {Neurological} {Disorders}. In: 2020 {IEEE} 22nd {International} {Workshop} on {Multimedia} {Signal} {Processing} ({MMSP}). pp.~1--6 (Sep 2020), iSSN: 2473-3628

\bibitem{Gross2012EffortShape}
Gross, M.M., Crane, E.A., Fredrickson, B.L.: Effort-{Shape} and kinematic assessment of bodily expression of emotion during gait. Human Movement Science  \textbf{31}(1),  202--221 (Feb 2012)

\bibitem{Gross2001CMU}
Gross, R., Shi, J.: The cmu motion of body (mobo) database. Tech. Rep. CMU-RI-TR-01-18, Carnegie Mellon University, Pittsburgh, PA (June 2001)

\bibitem{Grouvel2023dataset}
Grouvel, G., Carcreff, L., Moissenet, F., Armand, S.: A dataset of asymptomatic human gait and movements obtained from markers, imus, insoles and force plates. Scientific Data  \textbf{10}(1), ~180 (2023)

\bibitem{Han2006Individual}
Han, J., Bhanu, B.: Individual recognition using gait energy image. IEEE Transactions on Pattern Analysis and Machine Intelligence  \textbf{28}(2),  316--322 (2006)

\bibitem{Harris2022Survey}
Harris, E.J., Khoo, I.H., Demircan, E.: A {Survey} of {Human} {Gait}-{Based} {Artificial} {Intelligence} {Applications}. Frontiers in Robotics and AI  \textbf{8},  749274 (Jan 2022)

\bibitem{Hebenstreit2015Effect}
Hebenstreit, F., Leibold, A., Krinner, S., Welsch, G., Lochmann, M., Eskofier, B.M.: Effect of walking speed on gait sub phase durations. Human Movement Science  \textbf{43},  118--124 (Oct 2015)

\bibitem{hensley2020video}
Hensley, C.P., Millican, D., Hamilton, N., Yang, A., Lee, J., Chang, A.H.: Video-based motion analysis use: a national survey of orthopedic physical therapists. Physical therapy  \textbf{100}(10),  1759--1770 (2020)

\bibitem{Herath2017Going}
Herath, S., Harandi, M., Porikli, F.: Going deeper into action recognition: {A} survey. Image and Vision Computing  \textbf{60},  4--21 (Apr 2017)

\bibitem{Hirzer2011Person}
Hirzer, M., Beleznai, C., Roth, P.M., Bischof, H.: Person re-identification by descriptive and discriminative classification. In: Proceedings of the 17th {Scandinavian} conference on {Image} analysis. pp. 91--102. {SCIA}'11, Springer-Verlag, Berlin, Heidelberg (May 2011)

\bibitem{Hofmann2014TUM}
Hofmann, M., Geiger, J., Bachmann, S., Schuller, B., Rigoll, G.: The {TUM} {Gait} from {Audio}, {Image} and {Depth} ({GAID}) database: {Multimodal} recognition of subjects and traits. Journal of Visual Communication and Image Representation  \textbf{25}(1),  195--206 (2014)

\bibitem{Hofmann2011Gait}
Hofmann, M., Sural, S., Rigoll, G.: Gait {Recognition} in the {Presence} of {Occlusion}: {A} {New} {Dataset} and {Baseline} {Algorithms}. Václav Skala - UNION Agency (2011), accepted: 2014-03-18T09:37:28Z

\bibitem{Huang2008Action}
Huang, F., Xu, G.: Action recognition unrestricted by location and viewpoint variation. In: 2008 IEEE 8th International Conference on Computer and Information Technology Workshops. pp. 433--438 (2008). \doi{10.1109/CIT.2008.Workshops.41}

\bibitem{Iosa2012Effects}
Iosa, M., Fusco, A., Morone, G., Paolucci, S.: Effects of {Visual} {Deprivation} on {Gait} {Dynamic} {Stability}. The Scientific World Journal  \textbf{2012},  e974560 (May 2012), publisher: Hindawi

\bibitem{Ismail2017Torwads}
Ismail, A., Shouman, H., Cherry, A., Hajj-Hassan, M.: Torwads real time kinect anlysis system for early diagnosis of gait cycle abnormalities. In: 2017 {Fourth} {International} {Conference} on {Advances} in {Biomedical} {Engineering} ({ICABME}). pp.~1--4 (Oct 2017), iSSN: 2377-5696

\bibitem{Iwama2012OUISIR}
Iwama, H., Okumura, M., Makihara, Y., Yagi, Y.: The {OU}-{ISIR} {Gait} {Database} {Comprising} the {Large} {Population} {Dataset} and {Performance} {Evaluation} of {Gait} {Recognition}. IEEE Transactions on Information Forensics and Security  \textbf{7}(5),  1511--1521 (Oct 2012), conference Name: IEEE Transactions on Information Forensics and Security

\bibitem{Iwashita2010Person}
Iwashita, Y., Baba, R., Ogawara, K., Kurazume, R.: Person {Identification} from {Spatio}-temporal {3D} {Gait}. In: 2010 {International} {Conference} on {Emerging} {Security} {Technologies}. pp. 30--35. IEEE, Canterbury, TBD, United Kingdom (Sep 2010)

\bibitem{jahn2010gait}
Jahn, K., Zwergal, A., Schniepp, R.: Gait disturbances in old age: classification, diagnosis, and treatment from a neurological perspective. Deutsches {\"A}rzteblatt International  \textbf{107}(17), ~306 (2010)

\bibitem{Johnson2001multiview}
Johnson, A., Bobick, A.: A multi-view method for gait recognition using static body parameters. Lecture Notes in Computer Science (including subseries Lecture Notes in Artificial Intelligence and Lecture Notes in Bioinformatics)  \textbf{2091 LNCS},  301--311 (2001)

\bibitem{Jun2023Hybrid}
Jun, K., Lee, K., Lee, S., Lee, H., Kim, M.S.: Hybrid {Deep} {Neural} {Network} {Framework} {Combining} {Skeleton} and {Gait} {Features} for {Pathological} {Gait} {Recognition} (Apr 2023)

\bibitem{Jun2020Pathological}
Jun, K., Lee, Y., Lee, S., Lee, D.W., Kim, M.S.: Pathological {Gait} {Classification} {Using} {Kinect} v2 and {Gated} {Recurrent} {Neural} {Networks}. IEEE Access  \textbf{8},  139881--139891 (2020), conference Name: IEEE Access

\bibitem{Kastaniotis2016Posebased}
Kastaniotis, D., Theodorakopoulos, I., Fotopoulos, S.: Pose-based gait recognition with local gradient descriptors and hierarchically aggregated residuals. Journal of Electronic Imaging  \textbf{25}(6),  063019 (Dec 2016)

\bibitem{Kay2017Kinetics}
Kay, W., Carreira, J., Simonyan, K., Zhang, B., Hillier, C., Vijayanarasimhan, S., Viola, F., Green, T., Back, T., Natsev, P., Suleyman, M., Zisserman, A.: The {Kinetics} {Human} {Action} {Video} {Dataset} (May 2017), arXiv:1705.06950 [cs]

\bibitem{Keefe2014database}
Keefe, B.D., Villing, M., Racey, C., Strong, S.L., Wincenciak, J., Barraclough, N.E.: A database of whole-body action videos for the study of action, emotion, and untrustworthiness. Behavior Research Methods  \textbf{46}(4),  1042--1051 (Dec 2014)

\bibitem{Khokhlova2019Normal}
Khokhlova, M., Migniot, C., Morozov, A., Sushkova, O., Dipanda, A.: Normal and pathological gait classification lstm model. Artificial intelligence in medicine  \textbf{94},  54--66 (2019)

\bibitem{kidzinski2019automatic}
Kidzi{\'n}ski, {\L}., Delp, S., Schwartz, M.: Automatic real-time gait event detection in children using deep neural networks. PloS one  \textbf{14}(1),  e0211466 (2019)

\bibitem{kidzinski2020deep}
Kidzi{\'n}ski, {\L}., Yang, B., Hicks, J.L., Rajagopal, A., Delp, S.L., Schwartz, M.H.: Deep neural networks enable quantitative movement analysis using single-camera videos. Nature communications  \textbf{11}(1), ~4054 (2020)

\bibitem{Kim2022PathologicalGait}
Kim, J., Seo, H., Naseem, M.T., Lee, C.S.: Pathological-{Gait} {Recognition} {Using} {Spatiotemporal} {Graph} {Convolutional} {Networks} and {Attention} {Model}. Sensors (Basel, Switzerland)  \textbf{22}(13) (Jun 2022)

\bibitem{Kim2022deeplearning}
Kim, Y.K., Visscher, R.M.S., Viehweger, E., Singh, N.B., Taylor, W.R., Vogl, F.: A deep-learning approach for automatically detecting gait-events based on foot-marker kinematics in children with cerebral palsy-{Which} markers work best for which gait patterns? PloS One  \textbf{17}(10),  e0275878 (2022)

\bibitem{Kondragunta2020Gait}
Kondragunta, J., Hirtz, G.: Gait {Parameter} {Estimation} of {Elderly} {People} using {3D} {Human} {Pose} {Estimation} in {Early} {Detection} of {Dementia}. In: 2020 42nd {Annual} {International} {Conference} of the {IEEE} {Engineering} in {Medicine} \& {Biology} {Society} ({EMBC}). pp. 5798--5801 (Jul 2020), iSSN: 2694-0604

\bibitem{kong2022}
Kong, Y., Fu, Y.: Human action recognition and prediction: A survey. International Journal of Computer Vision  \textbf{130}(5),  1366--1401 (2022)

\bibitem{Konz2022STDeepGait}
Konz, L., Hill, A., Banaei-Kashani, F.: {ST}-{DeepGait}: {A} {Spatiotemporal} {Deep} {Learning} {Model} for {Human} {Gait} {Recognition}. Sensors  \textbf{22}(20), ~8075 (Jan 2022), number: 20 Publisher: Multidisciplinary Digital Publishing Institute

\bibitem{koonin2020trends}
Koonin, L.M., Hoots, B., Tsang, C.A., Leroy, Z., Farris, K., Jolly, B., Antall, P., McCabe, B., Zelis, C.B., Tong, I., et~al.: Trends in the use of telehealth during the emergence of the covid-19 pandemic—united states, january--march 2020. Morbidity and Mortality Weekly Report  \textbf{69}(43), ~1595 (2020)

\bibitem{Kozlow2018Gait}
Kozlow, P., Abid, N., Yanushkevich, S.: Gait {Type} {Analysis} {Using} {Dynamic} {Bayesian} {Networks}. Sensors  \textbf{18}(10), ~3329 (Oct 2018), number: 10 Publisher: Multidisciplinary Digital Publishing Institute

\bibitem{krebs1985reliability}
Krebs, D.E., Edelstein, J.E., Fishman, S.: Reliability of observational kinematic gait analysis. Physical Therapy  \textbf{65}(7),  1027--1033 (1985)

\bibitem{Kuehne2013HMDB51}
Kuehne, H., Jhuang, H., Stiefelhagen, R., Serre, T.: {HMDB51}: {A} {Large} {Video} {Database} for {Human} {Motion} {Recognition}. In: Nagel, W.E., Kröner, D.H., Resch, M.M. (eds.) High {Performance} {Computing} in {Science} and {Engineering} ‘12. pp. 571--582. Springer, Berlin, Heidelberg (2013)

\bibitem{Kuo2010Intercamera}
Kuo, C.H., Huang, C., Nevatia, R.: Inter-camera {Association} of {Multi}-target {Tracks} by {On}-{Line} {Learned} {Appearance} {Affinity} {Models}. In: Daniilidis, K., Maragos, P., Paragios, N. (eds.) Computer {Vision} – {ECCV} 2010. pp. 383--396. Lecture {Notes} in {Computer} {Science}, Springer, Berlin, Heidelberg (2010)

\bibitem{Kwolek2019Calibrated}
Kwolek, B., Michalczuk, A., Krzeszowski, T., Switonski, A., Josinski, H., Wojciechowski, K.: Calibrated and synchronized multi-view video and motion capture dataset for evaluation of gait recognition. Multimedia Tools and Applications  \textbf{78}(22),  32437--32465 (Nov 2019)

\bibitem{Laet2017Does}
Laet, T.D., Papageorgiou, E., Nieuwenhuys, A., Desloovere, K.: Does expert knowledge improve automatic probabilistic classification of gait joint motion patterns in children with cerebral palsy? PLOS ONE  \textbf{12}(6),  e0178378 (Jun 2017), publisher: Public Library of Science

\bibitem{Latorre2019Gait}
Latorre, J., Colomer, C., Alcañiz, M., Llorens, R.: Gait analysis with the {Kinect} v2: normative study with healthy individuals and comprehensive study of its sensitivity, validity, and reliability in individuals with stroke. Journal of NeuroEngineering and Rehabilitation  \textbf{16}(1), ~97 (Jul 2019)

\bibitem{le2022comprehensive}
Le, V.T., Tran-Trung, K., Hoang, V.T.: A comprehensive review of recent deep learning techniques for human activity recognition. Computational Intelligence and Neuroscience  \textbf{2022} (2022)

\bibitem{Leal-Taixe2015MOTChallenge}
Leal-Taixé, L., Milan, A., Reid, I., Roth, S., Schindler, K.: {MOTChallenge} 2015: {Towards} a {Benchmark} for {Multi}-{Target} {Tracking} (Apr 2015), arXiv:1504.01942 [cs]

\bibitem{Lee2019Abnormal}
Lee, D.W., Jun, K., Lee, S., Ko, J.K., Kim, M.S.: Abnormal {Gait} {Recognition} {Using} {3D} {Joint} information of {Multiple} {Kinects} {System} and {RNN}-{LSTM}. In: 2019 41st {Annual} {International} {Conference} of the {IEEE} {Engineering} in {Medicine} and {Biology} {Society} ({EMBC}). pp. 542--545 (Jul 2019), iSSN: 1558-4615

\bibitem{Lee2020Perceiving}
Lee, I.C., Pacheco, M.M., Lewek, M.D., Huang, H.: Perceiving amputee gait from biological motion: kinematics cues and effect of experience level. Scientific Reports  \textbf{10},  17093 (Oct 2020)

\bibitem{Lee2013Kinematic}
Lee, M., Kim, J., Son, J., Kim, Y.: Kinematic and kinetic analysis during forward and backward walking. Gait \& Posture  \textbf{38}(4),  674--678 (Sep 2013)

\bibitem{Li2018Unsupervised}
Li, M., Zhu, X., Gong, S.: Unsupervised {Person} {Re}-identification by {Deep} {Learning} {Tracklet} {Association} (Sep 2018), arXiv:1809.02874 [cs] version: 1

\bibitem{Li2023multimodala}
Li, N., Zhao, X.: A multi-modal dataset for gait recognition under occlusion. Applied Intelligence  \textbf{53}(2),  1517--1534 (Jan 2023)

\bibitem{Li2018Classification}
Li, Q., Wang, Y., Sharf, A., Cao, Y., Tu, C., Chen, B., Yu, S.: Classification of gait anomalies from kinect. The Visual Computer  \textbf{34}(2),  229--241 (Feb 2018)

\bibitem{Li2022MultiView}
Li, X., Makihara, Y., Xu, C., Yagi, Y.: Multi-{View} {Large} {Population} {Gait} {Database} {With} {Human} {Meshes} and {Its} {Performance} {Evaluation}. IEEE Transactions on Biometrics, Behavior, and Identity Science  \textbf{4}(2),  234--248 (Apr 2022), conference Name: IEEE Transactions on Biometrics, Behavior, and Identity Science

\bibitem{Liang2022GaitEdge}
Liang, J., Fan, C., Hou, S., Shen, C., Huang, Y., Yu, S.: Gaitedge: Beyond plain end-to-end gait recognition for better practicality. In: European Conference on Computer Vision. pp. 375--390. Springer (2022)

\bibitem{Liang2022reliability}
Liang, S., Zhang, Y., Diao, Y., Li, G., Zhao, G.: The reliability and validity of gait analysis system using {3D} markerless pose estimation algorithms. Frontiers in Bioengineering and Biotechnology  \textbf{10} (2022)

\bibitem{Liao2020modelbased}
Liao, R., Yu, S., An, W., Huang, Y.: A model-based gait recognition method with body pose and human prior knowledge. Pattern Recognition  \textbf{98},  107069 (Feb 2020)

\bibitem{Lin2020Gait}
Lin, B., Zhang, S., Bao, F.: Gait {Recognition} with {Multiple}-{Temporal}-{Scale} {3D} {Convolutional} {Neural} {Network}. In: Proceedings of the 28th {ACM} {International} {Conference} on {Multimedia}. pp. 3054--3062. {MM} '20, Association for Computing Machinery, New York, NY, USA (Oct 2020)

\bibitem{Little1998Recognizing}
Little, J., Boyd, J.: Recognizing people by their gait: The shape of motion. Videre  \textbf{1} (01 2001)

\bibitem{Liu2017PKUMMD}
Liu, C., Hu, Y., Li, Y., Song, S., Liu, J.: {PKU}-{MMD}: {A} {Large} {Scale} {Benchmark} for {Continuous} {Multi}-{Modal} {Human} {Action} {Understanding} (Mar 2017), arXiv:1703.07475 [cs]

\bibitem{Liu2020NTU}
Liu, J., Shahroudy, A., Perez, M., Wang, G., Duan, L.Y., Kot, A.C.: {NTU} {RGB}+{D} 120: {A} {Large}-{Scale} {Benchmark} for {3D} {Human} {Activity} {Understanding}. IEEE Transactions on Pattern Analysis and Machine Intelligence  \textbf{42}(10),  2684--2701 (Oct 2020), arXiv:1905.04757 [cs]

\bibitem{Liu2020PTBTIR}
Liu, Q., He, Z., Li, X., Zheng, Y.: {PTB}-{TIR}: {A} {Thermal} {Infrared} {Pedestrian} {Tracking} {Benchmark}. IEEE Transactions on Multimedia  \textbf{22}(3),  666--675 (Mar 2020), conference Name: IEEE Transactions on Multimedia

\bibitem{Lonini2022VideoBaseda}
Lonini, L., Moon, Y., Embry, K., Cotton, R.J., McKenzie, K., Jenz, S., Jayaraman, A.: Video-{Based} {Pose} {Estimation} for {Gait} {Analysis} in {Stroke} {Survivors} during {Clinical} {Assessments}: {A} {Proof}-of-{Concept} {Study}. Digital Biomarkers  \textbf{6}(1),  9--18 (Jan 2022)

\bibitem{Lopez-Fernandez2014AVA}
L{\'o}pez-Fern{\'a}ndez, D., Madrid-Cuevas, F.J., Carmona-Poyato, {\'A}., Mar{\'\i}n-Jim{\'e}nez, M.J., Mu{\~n}oz-Salinas, R.: The ava multi-view dataset for gait recognition. In: Activity Monitoring by Multiple Distributed Sensing: Second International Workshop, AMMDS 2014, Stockholm, Sweden, August 24, 2014, Revised Selected Papers 2. pp. 26--39. Springer (2014)

\bibitem{Loureiro2020Using}
Loureiro, J., Correia, P.L.: Using a skeleton gait energy image for pathological gait classification. In: 2020 15th IEEE International Conference on Automatic Face and Gesture Recognition (FG 2020). pp. 503--507 (2020)

\bibitem{lucyyardley2004psychosocial}
LUCYYARDLEY, M.J., HALLAM, R.S.: Psychosocial aspects of disorders affecting balance and gait. Clinical disorders of balance, posture and gait p.~360 (2004)

\bibitem{Mahyuddin2012Development}
Mahyuddin, A., Mihradi, S., Dirgantara, T., Moeliono, M., Prabowo, T.: Development of {Indonesian} gait database using {2D} optical motion analyzer system. ASEAN Engineering Journal Part A  \textbf{2},  62--72 (Dec 2012)

\bibitem{Makihara2012OUISIR}
Makihara, Y., Mannami, H., Tsuji, A., Hossain, M.A., Sugiura, K., Mori, A., Yagi, Y.: The {OU}-{ISIR} {Gait} {Database} {Comprising} the {Treadmill} {Dataset}. IPSJ Transactions on Computer Vision and Applications  \textbf{4},  53--62 (2012)

\bibitem{Mansur2014Gait}
Mansur, A., Makihara, Y., Aqmar, R., Yagi, Y.: Gait {Recognition} under {Speed} {Transition}. In: 2014 {IEEE} {Conference} on {Computer} {Vision} and {Pattern} {Recognition}. pp. 2521--2528. IEEE, Columbus, OH, USA (Jun 2014)

\bibitem{Masood2021Appearance}
Masood, H., Farooq, H.: An {Appearance} {Invariant} {Gait} {Recognition} {Technique} {Using} {Dynamic} {Gait} {Features}. International Journal of Optics  \textbf{2021},  1--15 (May 2021)

\bibitem{Matovski2010effect}
Matovski, D.S., Nixon, M.S., Mahmoodi, S., Carter, J.N.: The effect of time on the performance of gait biometrics. In: 2010 {Fourth} {IEEE} {International} {Conference} on {Biometrics}: {Theory}, {Applications} and {Systems} ({BTAS}). pp.~1--6 (Sep 2010)

\bibitem{mcbain2015impact}
McBain, H., Shipley, M., Newman, S.: The impact of self-monitoring in chronic illness on healthcare utilisation: a systematic review of reviews. BMC health services research  \textbf{15}(1),  1--10 (2015)

\bibitem{Mehdizadeh2022Toronto}
Mehdizadeh, S., Nabavi, H., Sabo, A., Arora, T., Iaboni, A., Taati, B.: The toronto older adults gait archive: video and 3d inertial motion capture data of older adults’ walking. Scientific data  \textbf{9}(1), ~398 (2022)

\bibitem{mengucc2014wearable}
Meng{\"u}{\c{c}}, Y., Park, Y.L., Pei, H., Vogt, D., Aubin, P.M., Winchell, E., Fluke, L., Stirling, L., Wood, R.J., Walsh, C.J.: Wearable soft sensing suit for human gait measurement. The International Journal of Robotics Research  \textbf{33}(14),  1748--1764 (2014)

\bibitem{Milan2016MOT16}
Milan, A., Leal-Taixe, L., Reid, I., Roth, S., Schindler, K.: {MOT16}: {A} {Benchmark} for {Multi}-{Object} {Tracking} (May 2016), arXiv:1603.00831 [cs]

\bibitem{Milstein2020Computer}
Milstein, A., Topol, E.J.: Computer vision's potential to improve health care. The Lancet  \textbf{395}(10236), ~1537 (May 2020), publisher: Elsevier

\bibitem{Moore2015elaborate}
Moore, J.K., Hnat, S.K., Van Den~Bogert, A.J.: An elaborate data set on human gait and the effect of mechanical perturbations. PeerJ  \textbf{3}, ~e918 (Apr 2015)

\bibitem{Mostayed2008Abnormal}
Mostayed, A., Mynuddin, M., Mazumder, G., Kim, S., Park, S.J.: Abnormal {Gait} {Detection} {Using} {Discrete} {Fourier} {Transform}. In: 2008 {International} {Conference} on {Multimedia} and {Ubiquitous} {Engineering} (mue 2008). pp. 36--40 (Apr 2008)

\bibitem{Mu2021ReSGait}
Mu, Z., Castro, F.M., Marín-Jiménez, M.J., Guil, N., Li, Y.R., Yu, S.: {ReSGait}: {The} {Real}-{Scene} {Gait} {Dataset}. In: 2021 {IEEE} {International} {Joint} {Conference} on {Biometrics} ({IJCB}). pp.~1--8 (Aug 2021), iSSN: 2474-9699

\bibitem{Muramatsu2012Arbitrary}
Muramatsu, D., Shiraishi, A., Makihara, Y., Yagi, Y.: Arbitrary view transformation model for gait person authentication. In: 2012 {IEEE} {Fifth} {International} {Conference} on {Biometrics}: {Theory}, {Applications} and {Systems} ({BTAS}). pp. 85--90 (Sep 2012)

\bibitem{Nakano2020Evaluation}
Nakano, N., Sakura, T., Ueda, K., Omura, L., Kimura, A., Iino, Y., Fukashiro, S., Yoshioka, S.: Evaluation of {3D} {Markerless} {Motion} {Capture} {Accuracy} {Using} {OpenPose} {With} {Multiple} {Video} {Cameras}. Frontiers in Sports and Active Living  \textbf{2}, ~50 (2020)

\bibitem{Needham2021accuracy}
Needham, L., Evans, M., Cosker, D.P., Wade, L., McGuigan, P.M., Bilzon, J.L., Colyer, S.L.: The accuracy of several pose estimation methods for 3d joint centre localisation. Scientific reports  \textbf{11}(1),  20673 (2021)

\bibitem{Nguyen2016SkeletonBased}
Nguyen, T.N., Huynh, H.H., Meunier, J.: Skeleton-{Based} {Abnormal} {Gait} {Detection}. Sensors  \textbf{16}(11), ~1792 (Nov 2016), number: 11 Publisher: Multidisciplinary Digital Publishing Institute

\bibitem{Nguyen2018Walking}
Nguyen, T.N., Meunier, J.: Walking gait dataset: Point clouds, skeletons, and silhouettes. Technical Report~1379, DIRO, University of Montreal (September 2018), available online: [link](https://www.iro.umontreal.ca/~labimage/GaitDataset/dataset.pdf)

\bibitem{Nieto-Hidalgo2016vision}
Nieto-Hidalgo, M., Ferrández-Pastor, F.J., Valdivieso-Sarabia, R.J., Mora-Pascual, J., García-Chamizo, J.M.: A vision based proposal for classification of normal and abnormal gait using {RGB} camera. Journal of Biomedical Informatics  \textbf{63},  82--89 (Oct 2016)

\bibitem{Nieto-Hidalgo2017Classification}
Nieto-Hidalgo, M., García-Chamizo, J.M.: Classification of {Pathologies} {Using} a {Vision} {Based} {Feature} {Extraction}. In: Ochoa, S.F., Singh, P., Bravo, J. (eds.) Ubiquitous {Computing} and {Ambient} {Intelligence}. pp. 265--274. Lecture {Notes} in {Computer} {Science}, Springer International Publishing, Cham (2017)

\bibitem{Nixon1999Automatic}
Nixon, M., Carter, J., Nash, J., Huang, P., Cunado, D., Stevenage, S.: Automatic gait recognition. In: {IEE} {Colloquium} on {Motion} {Analysis} and {Tracking} ({Ref}. {No}. 1999/103). pp. 3/1--3/6 (May 1999)

\bibitem{Nunes2019GRIDDS}
Nunes, J.F., Moreira, P.M., Tavares, J.M.R.S.: {GRIDDS} - {A} {Gait} {Recognition} {Image} and {Depth} {Dataset}. In: Tavares, J.M.R.S., Natal~Jorge, R.M. (eds.) {VipIMAGE} 2019. pp. 343--352. Lecture {Notes} in {Computational} {Vision} and {Biomechanics}, Springer International Publishing, Cham (2019)

\bibitem{Ogale2007ViewInvariant}
Ogale, A.S., Karapurkar, A., Aloimonos, Y.: View-{Invariant} {Modeling} and {Recognition} of {Human} {Actions} {Using} {Grammars}. In: Vidal, R., Heyden, A., Ma, Y. (eds.) Dynamical {Vision}, vol.~4358, pp. 115--126. Springer Berlin Heidelberg, Berlin, Heidelberg (2007), series Title: Lecture Notes in Computer Science

\bibitem{DeOliveiraSilva2020Gait}
de~Oliveira~Silva, F., Ferreira, J.V., Plácido, J., Chagas, D., Praxedes, J., Guimarães, C., Batista, L.A., Laks, J., Deslandes, A.C.: Gait analysis with videogrammetry can differentiate healthy elderly, mild cognitive impairment, and {Alzheimer}'s disease: {A} cross-sectional study. Experimental Gerontology  \textbf{131},  110816 (Mar 2020)

\bibitem{OMalley1997Fuzzy}
O'Malley, M., Abel, M., Damiano, D., Vaughan, C.: Fuzzy clustering of children with cerebral palsy based on temporal- distance gait parameters. IEEE Transactions on Rehabilitation Engineering  \textbf{5}(4),  300--309 (1997)

\bibitem{Ortells2018Visionbased}
Ortells, J., Herrero-Ezquerro, M.T., Mollineda, R.A.: Vision-based gait impairment analysis for aided diagnosis. Medical \& Biological Engineering \& Computing  \textbf{56}(9),  1553--1564 (Sep 2018)

\bibitem{Pachon-Suescun2020Abnormal}
Pachon-Suescun, C.G., Pinzon-Arenas, J.O., Jimenez-Moreno, R.: Abnormal gait detection by means of {LSTM}. International Journal of Electrical and Computer Engineering (IJECE)  \textbf{10}(2), ~1495 (Apr 2020)

\bibitem{Parashar2022Intraclass}
Parashar, A., Shekhawat, R.S., Ding, W., Rida, I.: Intra-class variations with deep learning-based gait analysis: {A} comprehensive survey of covariates and methods. Neurocomputing  \textbf{505},  315--338 (Sep 2022)

\bibitem{Perera2018Human}
Perera, A.G., Law, Y.W., Chahl, J.: Human {Pose} and {Path} {Estimation} from {Aerial} {Video} {Using} {Dynamic} {Classifier} {Selection}. Cognitive Computation  \textbf{10}(6),  1019--1041 (Dec 2018)

\bibitem{Prochazka2015Bayesian}
Procházka, A., Vyšata, O., Vališ, M., Ťupa, O., Schätz, M., Mařík, V.: Bayesian classification and analysis of gait disorders using image and depth sensors of {Microsoft} {Kinect}. Digital Signal Processing  \textbf{47},  169--177 (Dec 2015)

\bibitem{Qu2011Effects}
Qu, X., Yeo, J.C.: Effects of load carriage and fatigue on gait characteristics. Journal of Biomechanics  \textbf{44}(7),  1259--1263 (Apr 2011)

\bibitem{Rahmani2016Histogram}
Rahmani, H., Mahmood, A., Huynh, D., Mian, A.: Histogram of {Oriented} {Principal} {Components} for {Cross}-{View} {Action} {Recognition}. IEEE Transactions on Pattern Analysis and Machine Intelligence  \textbf{38}(12),  2430--2443 (Dec 2016), conference Name: IEEE Transactions on Pattern Analysis and Machine Intelligence

\bibitem{Rana20213D}
Rana, S.P., Dey, M., Ghavami, M., Dudley, S.: 3-{D} {Gait} {Abnormality} {Detection} {Employing} {Contactless} {IR}-{UWB} {Sensing} {Phenomenon}. IEEE Transactions on Instrumentation and Measurement  \textbf{70},  1--10 (2021), conference Name: IEEE Transactions on Instrumentation and Measurement

\bibitem{Ristani2016Performance}
Ristani, E., Solera, F., Zou, R.S., Cucchiara, R., Tomasi, C.: Performance {Measures} and a {Data} {Set} for {Multi}-{Target}, {Multi}-{Camera} {Tracking} (Sep 2016), arXiv:1609.01775 [cs]

\bibitem{roberts2017biomechanical}
Roberts, M., Mongeon, D., Prince, F.: Biomechanical parameters for gait analysis: a systematic review of healthy human gait. Phys. Ther. Rehabil  \textbf{4}(6) (2017)

\bibitem{Rocha2018System}
Rocha, A.P., Choupina, H.M.P., Vilas-Boas, M.d.C., Fernandes, J.M., Cunha, J.P.S.: System for automatic gait analysis based on a single {RGB}-{D} camera. PLOS ONE  \textbf{13}(8),  e0201728 (Aug 2018), publisher: Public Library of Science

\bibitem{FilipiGoncalvesDosSantos2023Gait}
Filipi Gonçalves~dos Santos, C., Oliveira, D.d.S., A.~Passos, L., Gonçalves~Pires, R., Felipe Silva~Santos, D., Pascotti~Valem, L., P.~Moreira, T., Cleison S.~Santana, M., Roder, M., Paulo~Papa, J., Colombo, D.: Gait {Recognition} {Based} on {Deep} {Learning}: {A} {Survey}. ACM Computing Surveys  \textbf{55}(2),  1--34 (Feb 2023)

\bibitem{Sardari2020VINet}
Sardari, F., Paiement, A., Hannuna, S., Mirmehdi, M.: {VI}-{Net}—{View}-{Invariant} {Quality} of {Human} {Movement} {Assessment}. Sensors  \textbf{20}(18), ~5258 (Jan 2020), number: 18 Publisher: Multidisciplinary Digital Publishing Institute

\bibitem{Sardari2019ViewInvariant}
Sardari, F., Paiement, A., Mirmehdi, M.: View-{Invariant} {Pose} {Analysis} for {Human} {Movement} {Assessment} from {RGB} {Data}. In: Ricci, E., Rota~Bulò, S., Snoek, C., Lanz, O., Messelodi, S., Sebe, N. (eds.) Image {Analysis} and {Processing} – {ICIAP} 2019. pp. 237--248. Lecture {Notes} in {Computer} {Science}, Springer International Publishing, Cham (2019)

\bibitem{Sarkar2005humanID}
Sarkar, S., Phillips, P., Liu, Z., Vega, I., Grother, P., Bowyer, K.: The {humanID} gait challenge problem: data sets, performance, and analysis. IEEE Transactions on Pattern Analysis and Machine Intelligence  \textbf{27}(2),  162--177 (Feb 2005), conference Name: IEEE Transactions on Pattern Analysis and Machine Intelligence

\bibitem{Schuldt2004Recognizing}
Schuldt, C., Laptev, I., Caputo, B.: Recognizing human actions: a local {SVM} approach. In: Proceedings of the 17th {International} {Conference} on {Pattern} {Recognition}, 2004. {ICPR} 2004. vol.~3, pp. 32--36 Vol.3 (Aug 2004)

\bibitem{Seifallahi2020Alzheimer}
Seifallahi, M., Soltanizadeh, H., Mehraban, A.: Alzheimer’s disease detection using skeleton data recorded with {Kinect} camera. Cluster Computing  \textbf{23} (Jun 2020)

\bibitem{Sepas-Moghaddam2022Deep}
Sepas-Moghaddam, A., Etemad, A.: Deep {Gait} {Recognition}: {A} {Survey} (Feb 2022), arXiv:2102.09546 [cs]

\bibitem{Sepas-Moghaddam2023}
Sepas-Moghaddam, A., Etemad, A.: Deep gait recognition: A survey. IEEE Transactions on Pattern Analysis and Machine Intelligence  \textbf{45}(1),  264--284 (2023)

\bibitem{Sethi2022comprehensive}
Sethi, D., Bharti, S., Prakash, C.: A comprehensive survey on gait analysis: History, parameters, approaches, pose estimation, and future work. Artificial Intelligence in Medicine  \textbf{129},  102314 (2022)

\bibitem{sethi2022}
Sethi, D., Prakash, C., Bharti, S.: Latest trends in gait analysis using deep learning techniques: A systematic review. In: Artificial Intelligence and Speech Technology: Third International Conference, AIST 2021, Delhi, India, November 12--13, 2021, Revised Selected Papers. pp. 363--375. Springer (2022)

\bibitem{Shahroudy2016NTU}
Shahroudy, A., Liu, J., Ng, T.T., Wang, G.: {NTU} {RGB}+{D}: {A} {Large} {Scale} {Dataset} for {3D} {Human} {Activity} {Analysis}. In: 2016 {IEEE} {Conference} on {Computer} {Vision} and {Pattern} {Recognition} ({CVPR}). pp. 1010--1019 (Jun 2016), iSSN: 1063-6919

\bibitem{Shao2020FineGym}
Shao, D., Zhao, Y., Dai, B., Lin, D.: Finegym: A hierarchical video dataset for fine-grained action understanding. In: Proceedings of the IEEE/CVF conference on computer vision and pattern recognition. pp. 2616--2625 (2020)

\bibitem{sheng2021multi}
Sheng, W., Li, X.: Multi-task learning for gait-based identity recognition and emotion recognition using attention enhanced temporal graph convolutional network. Pattern Recognition  \textbf{114},  107868 (2021)

\bibitem{Shiraga2016GEINet}
Shiraga, K., Makihara, Y., Muramatsu, D., Echigo, T., Yagi, Y.: {GEINet}: {View}-invariant gait recognition using a convolutional neural network. In: 2016 {International} {Conference} on {Biometrics} ({ICB}). pp.~1--8. IEEE, Halmstad, Sweden (Jun 2016)

\bibitem{Shutler2004Large}
Shutler, J.D., Grant, M.G., Nixon, M.S., Carter, J.N.: On a {Large} {Sequence}-{Based} {Human} {Gait} {Database}. In: Lotfi, A., Garibaldi, J.M. (eds.) Applications and {Science} in {Soft} {Computing}. pp. 339--346. Advances in {Soft} {Computing}, Springer, Berlin, Heidelberg (2004)

\bibitem{sicard2002gait}
Sicard-Rosenbaum, L., Light, K.E., Behrman, A.L.: Gait, lower extremity strength, and self-assessed mobility after hip arthroplasty. The Journals of Gerontology Series A: Biological Sciences and Medical Sciences  \textbf{57}(1),  M47--M51 (2002)

\bibitem{Sigal2010HumanEva}
Sigal, L., Balan, A.O., Black, M.J.: {HumanEva}: {Synchronized} {Video} and {Motion} {Capture} {Dataset} and {Baseline} {Algorithm} for {Evaluation} of {Articulated} {Human} {Motion}. International Journal of Computer Vision  \textbf{87}(1),  4--27 (Mar 2010)

\bibitem{Simonyan2014TwoStream}
Simonyan, K., Zisserman, A.: Two-{Stream} {Convolutional} {Networks} for {Action} {Recognition} in {Videos} (Nov 2014), arXiv:1406.2199 [cs]

\bibitem{Singh2019MultiGait}
Singh, J.P., Arora, S., Jain, S., Singh~SoM, U.P.: A multi-gait dataset for human recognition under occlusion scenario. In: 2019 International Conference on Issues and Challenges in Intelligent Computing Techniques (ICICT). vol.~1, pp.~1--6 (2019). \doi{10.1109/ICICT46931.2019.8977673}

\bibitem{Singh2010MuHAVi}
Singh, S., Velastin, S.A., Ragheb, H.: {MuHAVi}: {A} {Multicamera} {Human} {Action} {Video} {Dataset} for the {Evaluation} of {Action} {Recognition} {Methods}. In: 2010 7th {IEEE} {International} {Conference} on {Advanced} {Video} and {Signal} {Based} {Surveillance}. pp. 48--55 (Aug 2010)

\bibitem{Soltaninejad2019KinFOG}
Soltaninejad, S., Cheng, I., Basu, A.: Kin-{FOG}: {Automatic} {Simulated} {Freezing} of {Gait} ({FOG}) {Assessment} {System} for {Parkinson}’s {Disease}. Sensors  \textbf{19}(10), ~2416 (Jan 2019), number: 10 Publisher: Multidisciplinary Digital Publishing Institute

\bibitem{Song2023CASIAE}
Song, C., Huang, Y., Wang, W., Wang, L.: {CASIA}-{E}: {A} {Large} {Comprehensive} {Dataset} for {Gait} {Recognition}. IEEE Transactions on Pattern Analysis and Machine Intelligence  \textbf{45}(3),  2801--2815 (2023)

\bibitem{Song2019GaitNet}
Song, C., Huang, Y., Huang, Y., Jia, N., Wang, L.: {GaitNet}: {An} end-to-end network for gait based human identification. Pattern Recognition  \textbf{96},  106988 (Dec 2019)

\bibitem{Song2017Regionbased}
Song, G., Leng, B., Liu, Y., Hetang, C., Cai, S.: Region-based {Quality} {Estimation} {Network} for {Large}-scale {Person} {Re}-identification (Dec 2017), arXiv:1711.08766 [cs] version: 2

\bibitem{Stenum2021Twodimensional}
Stenum, J., Rossi, C., Roemmich, R.T.: Two-dimensional video-based analysis of human gait using pose estimation. PLOS Computational Biology  \textbf{17}(4),  e1008935 (Apr 2021), publisher: Public Library of Science

\bibitem{Stergiou2020Biomechanics}
Stergiou, N.: Biomechanics and gait analysis. Academic Press, London ; San Diego, CA (2020), oCLC: on1151917798

\bibitem{Takemura2018Multiview}
Takemura, N., Makihara, Y., Muramatsu, D., Echigo, T., Yagi, Y.: Multi-view large population gait dataset and its performance evaluation for cross-view gait recognition. IPSJ Transactions on Computer Vision and Applications  \textbf{10}(1), ~4 (Feb 2018)

\bibitem{Tan2006Efficient}
Tan, D., Huang, K., Yu, S., Tan, T.: Efficient night gait recognition based on template matching. In: 18th International Conference on Pattern Recognition (ICPR'06). vol.~3, pp. 1000--1003 (2006). \doi{10.1109/ICPR.2006.478}

\bibitem{Tanawongsuwan2004Modelling}
Tanawongsuwan, R., Bobick, A.: Modelling the effects of walking speed on appearance-based gait recognition. In: Proceedings of the 2004 IEEE Computer Society Conference on Computer Vision and Pattern Recognition, 2004. CVPR 2004. vol.~2, pp. II--II (2004). \doi{10.1109/CVPR.2004.1315244}

\bibitem{Tao2016comparative}
Tao, L., Paiement, A., Damen, D., Mirmehdi, M., Hannuna, S., Camplani, M., Burghardt, T., Craddock, I.: A comparative study of pose representation and dynamics modelling for online motion quality assessment. Computer Vision and Image Understanding  \textbf{148},  136--152 (Jul 2016)

\bibitem{Teepe2022Deeper}
Teepe, T., Gilg, J., Herzog, F., Hörmann, S., Rigoll, G.: Towards a {Deeper} {Understanding} of {Skeleton}-based {Gait} {Recognition} (Apr 2022), arXiv:2204.07855 [cs]

\bibitem{Tian2022Skeletonbased}
Tian, H., Ma, X., Wu, H., Li, Y.: Skeleton-based abnormal gait recognition with spatio-temporal attention enhanced gait-structural graph convolutional networks. Neurocomputing  \textbf{473},  116--126 (Feb 2022)

\bibitem{Topham2023diverse}
Topham, L.K., Khan, W., Al-Jumeily, D., Waraich, A., Hussain, A.J.: A diverse and multi-modal gait dataset of indoor and outdoor walks acquired using multiple cameras and sensors. Scientific Data  \textbf{10}(1), ~320 (2023)

\bibitem{Toro2003review}
Toro, B., Nester, C., Farren, P.: A review of observational gait assessment in clinical practice. Physiotherapy Theory and Practice  \textbf{19}(3),  137--149 (Jan 2003)

\bibitem{totah2019impact}
Totah, D., Menon, M., Jones-Hershinow, C., Barton, K., Gates, D.H.: The impact of ankle-foot orthosis stiffness on gait: A systematic literature review. Gait \& posture  \textbf{69},  101--111 (2019)

\bibitem{Tsukagoshi2020Noninvasive}
Tsukagoshi, S., Furuta, M., Hirayanagi, K., Furuta, N., Nakazato, S., Fujii, M., Yuminaka, Y., Ikeda, Y.: Noninvasive and quantitative evaluation of movement disorder disability using an infrared depth sensor. Journal of Clinical Neuroscience  \textbf{71},  135--140 (Jan 2020)

\bibitem{Uddin2018OUISIR}
Uddin, M.Z., Ngo, T.T., Makihara, Y., Takemura, N., Li, X., Muramatsu, D., Yagi, Y.: The {OU}-{ISIR} {Large} {Population} {Gait} {Database} with real-life carried object and its performance evaluation. IPSJ Transactions on Computer Vision and Applications  \textbf{10}(1), ~5 (May 2018)

\bibitem{Vafadar2022Assessment}
Vafadar, S., Skalli, W., Bonnet-Lebrun, A., Assi, A., Gajny, L.: Assessment of a novel deep learning-based marker-less motion capture system for gait study. Gait \& Posture  \textbf{94},  138--143 (May 2022)

\bibitem{Vafadar2021novel}
Vafadar, S., Skalli, W., Bonnet-Lebrun, A., Khalif{\'e}, M., Renaudin, M., Hamza, A., Gajny, L.: A novel dataset and deep learning-based approach for marker-less motion capture during gait. Gait \& Posture  \textbf{86},  70--76 (2021)

\bibitem{VanCriekinge2023fullbody}
Van~Criekinge, T., Saeys, W., Truijen, S., Vereeck, L., Sloot, L.H., Hallemans, A.: A full-body motion capture gait dataset of 138 able-bodied adults across the life span and 50 stroke survivors. Scientific Data  \textbf{10}(1), ~852 (Dec 2023), number: 1 Publisher: Nature Publishing Group

\bibitem{Venture2014Recognizing}
Venture, G., Kadone, H., Zhang, T., Grèzes, J., Berthoz, A., Hicheur, H.: Recognizing {Emotions} {Conveyed} by {Human} {Gait}. International Journal of Social Robotics  \textbf{6}(4),  621--632 (Nov 2014)

\bibitem{Vilas-Boas2019Validation}
Vilas-Boas, M.d.C., Rocha, A.P., Choupina, H.M.P., Cardoso, M.N., Fernandes, J.M., Coelho, T., Cunha, J.P.S.: Validation of a {Single} {RGB}-{D} {Camera} for {Gait} {Assessment} of {Polyneuropathy} {Patients}. Sensors  \textbf{19}(22), ~4929 (Jan 2019), number: 22 Publisher: Multidisciplinary Digital Publishing Institute

\bibitem{Wang2024VideoBased}
Wang, D., Zouaoui, C., Jang, J., Drira, H., Seo, H.: Video-{Based} {Gait} {Analysis} for {Assessing} {Alzheimer}’s {Disease} and {Dementia} with {Lewy} {Bodies}. In: Wu, S., Shabestari, B., Xing, L. (eds.) Applications of Medical Artificial Intelligence. vol. 14313, pp. 72--82. Springer Nature Switzerland, Cham (2024)

\bibitem{Wang2014Crossview}
wang, J., Nie, X., Xia, Y., Wu, Y., Zhu, S.C.: Cross-view {Action} {Modeling}, {Learning} and {Recognition} (May 2014), arXiv:1405.2941 [cs]

\bibitem{Wang2003Silhouette}
Wang, L., Tan, T., Ning, H., Hu, W.: Silhouette {Analysis}-{Based} {Gait} {Recognition} for {Human} {Identification}. IEEE Transactions on Pattern Analysis and Machine Intelligence  \textbf{25}(12),  1505--1518 (2003)

\bibitem{Wang2016Temporal}
Wang, L., Xiong, Y., Wang, Z., Qiao, Y., Lin, D., Tang, X., Van~Gool, L.: Temporal {Segment} {Networks}: {Towards} {Good} {Practices} for {Deep} {Action} {Recognition}. In: Leibe, B., Matas, J., Sebe, N., Welling, M. (eds.) Computer {Vision} – {ECCV} 2016, vol.~9912, pp. 20--36. Springer International Publishing, Cham (2016)

\bibitem{Wang2021Gait}
Wang, T., Li, C., Wu, C., Zhao, C., Sun, J., Peng, H., Hu, X., Hu, B.: A {Gait} {Assessment} {Framework} for {Depression} {Detection} {Using} {Kinect} {Sensors}. IEEE Sensors Journal  \textbf{21}(3),  3260--3270 (Feb 2021)

\bibitem{Weinland2006Free}
Weinland, D., Ronfard, R., Boyer, E.: Free viewpoint action recognition using motion history volumes. Computer Vision and Image Understanding  \textbf{104}(2),  249--257 (Nov 2006)

\bibitem{whittle2014gait}
Whittle, M.W.: Gait analysis: an introduction. Butterworth-Heinemann (2014)

\bibitem{wiles2003use}
Wiles, C., Newcombe, R., Fuller, K., Jones, A., Price, M.: Use of videotape to assess mobility in a controlled randomized crossover trial of physiotherapy in chronic multiple sclerosis. Clinical rehabilitation  \textbf{17}(3),  256--263 (2003)

\bibitem{Wolf2006Automated}
Wolf, S., Loose, T., Schablowski, M., Döderlein, L., Rupp, R., Gerner, H.J., Bretthauer, G., Mikut, R.: Automated feature assessment in instrumented gait analysis. Gait \& Posture  \textbf{23}(3),  331--338 (Apr 2006)

\bibitem{Wren2020Clinical}
Wren, T.A.L., Tucker, C.A., Rethlefsen, S.A., Gorton, G.E., Õunpuu, S.: Clinical efficacy of instrumented gait analysis: {Systematic} review 2020 update. Gait \& Posture  \textbf{80},  274--279 (Jul 2020)

\bibitem{wren2011efficacy}
Wren, T.A., Gorton~III, G.E., Ounpuu, S., Tucker, C.A.: Efficacy of clinical gait analysis: A systematic review. Gait \& posture  \textbf{34}(2),  149--153 (2011)

\bibitem{Xu2017OUISIR}
Xu, C., Makihara, Y., Ogi, G., Li, X., Yagi, Y., Lu, J.: The {OU}-{ISIR} {Gait} {Database} comprising the {Large} {Population} {Dataset} with {Age} and performance evaluation of age estimation. IPSJ Transactions on Computer Vision and Applications  \textbf{9}(1), ~24 (Dec 2017)

\bibitem{Xu2019Gaitbased}
Xu, C., Makihara, Y., Yagi, Y., Lu, J.: Gait-based age progression/regression: a baseline and performance evaluation by age group classification and cross-age gait identification. Machine Vision and Applications  \textbf{30}(4),  629--644 (Jun 2019)

\bibitem{Yamada2020Gaitbased}
Yamada, H., Ahn, J., Mozos, O.M., Iwashita, Y., Kurazume, R.: Gait-based person identification using {3D} {LiDAR} and long short-term memory deep networks. Advanced Robotics  \textbf{34}(18),  1201--1211 (Sep 2020)

\bibitem{Yamamoto2022Verification}
Yamamoto, M., Shimatani, K., Ishige, Y., Takemura, H.: Verification of gait analysis method fusing camera-based pose estimation and an imu sensor in various gait conditions. Scientific reports  \textbf{12}(1),  17719 (2022)

\bibitem{Yang2022Data}
Yang, J., Lu, H., Li, C., Hu, X., Hu, B.: Data augmentation for depression detection using skeleton-based gait information. Medical \& Biological Engineering \& Computing  \textbf{60}(9),  2665--2679 (2022)

\bibitem{Yousef2023Modelbased}
Yousef, R.N., Khalil, A.T., Samra, A.S., Ata, M.M.: Model-based and model-free deep features fusion for high performed human gait recognition. The Journal of Supercomputing  (Mar 2023)

\bibitem{Yu2006framework}
Yu, S., Tan, D., Tan, T.: A framework for evaluating the effect of view angle, clothing and carrying condition on gait recognition. In: 18th International Conference on Pattern Recognition (ICPR'06). vol.~4, pp. 441--444 (2006), iSSN: 1051-4651

\bibitem{Yu2021Gaitbased}
Yu, S., Li, X.: Gait-based {Emotion} {Recognition} {Using} {Spatial} {Temporal} {Graph} {Convolutional} {Networks}. In: 2021 {International} {Conference} on {Computer} {Information} {Science} and {Artificial} {Intelligence} ({CISAI}). pp. 190--193 (Sep 2021)

\bibitem{Yun2014Statistical}
Yun, Y., Kim, H.C., Shin, S.Y., Lee, J., Deshpande, A.D., Kim, C.: Statistical method for prediction of gait kinematics with {Gaussian} process regression. Journal of Biomechanics  \textbf{47}(1),  186--192 (Jan 2014)

\bibitem{Zago20203D}
Zago, M., Luzzago, M., Marangoni, T., De~Cecco, M., Tarabini, M., Galli, M.: {3D} {Tracking} of {Human} {Motion} {Using} {Visual} {Skeletonization} and {Stereoscopic} {Vision}. Frontiers in Bioengineering and Biotechnology  \textbf{8}, ~181 (2020)

\bibitem{Zhang2013Estimation}
Zhang, D., Wang, Y., Zhang, Z., Hu, M.: Estimation of view angles for gait using a robust regression method. Multimedia Tools and Applications  \textbf{65}(3),  419--439 (Aug 2013)

\bibitem{Zhang2023LargeScale}
Zhang, P., Dou, H., Zhang, W., Zhao, Y., Qin, Z., Hu, D., Fang, Y., Li, X.: A {Large}-{Scale} {Synthetic} {Gait} {Dataset} {Towards} in-the-{Wild} {Simulation} and {Comparison} {Study}. ACM Transactions on Multimedia Computing, Communications and Applications  \textbf{19}(1) (2023)

\bibitem{Zhang2023RealGait}
Zhang, S., Wang, Y., Chai, T., Li, A., Jain, A.K.: {RealGait}: {Gait} {Recognition} for {Person} {Re}-{Identification} (Feb 2023), arXiv:2201.04806 [cs]

\bibitem{Zhang2019Gait}
Zhang, Z., Tran, L., Yin, X., Atoum, Y., Liu, X., Wan, J., Wang, N.: Gait recognition via disentangled representation learning. In: 2019 IEEE/CVF Conference on Computer Vision and Pattern Recognition (CVPR). pp. 4705--4714 (2019). \doi{10.1109/CVPR.2019.00484}

\bibitem{Zheng2022Gait}
Zheng, J., Liu, X., Liu, W., He, L., Yan, C., Mei, T.: Gait recognition in the wild with dense 3d representations and a benchmark. In: Proceedings of the IEEE/CVF Conference on Computer Vision and Pattern Recognition. pp. 20228--20237 (2022)

\bibitem{Zheng2016MARS}
Zheng, L., Bie, Z., Sun, Y., Wang, J., Su, C., Wang, S., Tian, Q.: {MARS}: {A} {Video} {Benchmark} for {Large}-{Scale} {Person} {Re}-{Identification}. In: Computer {Vision} – {ECCV} 2016. pp. 868--884. Springer, Cham (2016), iSSN: 1611-3349

\bibitem{Zhou2024Portable}
Zhou, C., Feng, D., Chen, S., Ban, N., Pan, J.: Portable vision-based gait assessment for post-stroke rehabilitation using an attention-based lightweight {CNN}. Expert Systems with Applications  \textbf{238},  122074 (Mar 2024)

\bibitem{Zhu2022Gait}
Zhu, Z., Guo, X., Yang, T., Huang, J., Deng, J., Huang, G., Du, D., Lu, J., Zhou, J.: Gait {Recognition} in the {Wild}: {A} {Benchmark} (May 2022), arXiv:2205.02692 [cs]

\end{thebibliography}
\end{document}